\documentclass[a4paper, 11pt]{article}
\pdfoutput=1  
\usepackage{etoolbox,verbatim}
\usepackage{wrapfig,pifont}

\usepackage[table]{xcolor}
\usepackage{algorithm}
\usepackage{algorithmic}

\newtoggle{arxiv}
\toggletrue{arxiv} 

\newtoggle{customthms}
\toggletrue{customthms} 

\newtoggle{colt}
\togglefalse{colt} 

\newcommand{\DAT}{\textcolor{primalcolor}{\textbf{\texttt{DAT}}}}
\newcommand{\HHG}{\textcolor{lightblue}{\textbf{\texttt{HG-H}}}}

\usepackage{xcolor}
\usepackage{tcolorbox}
\tcbuselibrary{skins,breakable}

\tcbset{
  aibox/.style={
      width=\linewidth,
      top=6pt,
      bottom=0pt,
      left=1.5pt,
      right=1.5pt,
      colback=blue!60!gray!5,
      colframe=black,
      colbacktitle=black,
      enhanced,
      center,
      attach boxed title to top left={yshift=-0.1in,xshift=0.15in},
      boxed title style={boxrule=0pt,colframe=white,},
    }
}
\newtcolorbox{AIbox}[2][]{aibox,title=#2,#1}

\definecolor{primalcolor}{HTML}{A60000}
\definecolor{contrarycolor}{HTML}{00A6A6}
\definecolor{darkcontrarycolor}{HTML}{004C4C}
\definecolor{lightblue}{HTML}{2970CC}
\definecolor{lightpurple}{HTML}{673147}
\definecolor{ForestGreen}{HTML}{FF5733}
\definecolor{myred}{HTML}{AA4A44}
\definecolor{hyppurple}{HTML}{800080}

\newcommand{\linkcolor}{darkcontrarycolor}
\newcommand{\urlcolor}{darkcontrarycolor}
\newcommand{\citecolor}{darkcontrarycolor}

\newcommand{\thmcolordark}{red!30!black}

\usepackage[T1]{fontenc}
\usepackage{capt-of}
\usepackage[font=small,labelfont=bf]{caption}

\usepackage{natbib}
\usepackage{scalefnt}
\usepackage{amssymb,upgreek,bm}
\usepackage{mathtools}
\usepackage[english]{babel}  
\usepackage{multirow,tcolorbox,tikz-cd,turnstile,xspace,xparse}
\usepackage{relsize,breakcites,appendix}
\usepackage{longtable,tablefootnote,booktabs,float,makecell}
\usepackage[normalem]{ulem}
\usepackage{tabu}
\usepackage[shortlabels]{enumitem} 
\usepackage{footnote}
\usepackage{boxedminipage}
\usepackage{multirow,nicefrac}
\usepackage{verbatim} 

\usepackage[colorlinks=true,linkcolor=\linkcolor,urlcolor=\urlcolor,citecolor=\citecolor,breaklinks]{hyperref}

\usepackage{prettyref}

\usepackage[capitalise,nameinlink]{cleveref}

\iftoggle{colt}{
    \DeclareRobustCommand{\qed}{
        \usepackage{thmtools}
          \ifmmode \mathqed
          \else
            \leavevmode\unskip\penalty9999 \hbox{}\nobreak\hfill
            \quad\hbox{\qedsymbol}%
          \fi
    }
}
{
    \usepackage{amsthm}
     
}
\usepackage{thm-restate}

\iftoggle{arxiv}
{
    \usepackage{mathrsfs}
    \usepackage[
        a4paper,
        top=1in,
        bottom=1in,
        left=1in,
        right=1in]{geometry}
    \usepackage[bitstream-charter]{mathdesign}
     
    \usepackage[scaled=0.92]{PTSans}

    \usepackage{subfig}

    \usepackage{fullpage}

}
{
    \usepackage{mathrsfs}
}

\DeclareMathAlphabet{\mathbfsf}{\encodingdefault}{\sfdefault}{bx}{n}

\numberwithin{equation}{section}

\iftoggle{arxiv}
{
    }
{
    
}
\iftoggle{arxiv}
{
    
}
{
    
}

\usepackage{thmtools}

\Crefname{equation}{Eq.}{Eqs.}
\Crefname{assumption}{Assumption}{Assumptions}
\Crefname{condition}{Condition}{Conditions}
\Crefname{claim}{Claim}{Claims}
\Crefname{property}{Property}{Properties}
\Crefname{construction}{Construction}{Constructions}

\declaretheoremstyle[
    headformat=\normalfont\textcolor{\thmcolordark}{\bfseries\NAME\,\NUMBER}\NOTE,%
    notefont={\normalfont\textcolor{\thmcolordark}{\bfseries}}, 
    notebraces={}{},
    bodyfont=\normalfont\itshape,
    spaceabove = 6pt,
    spacebelow = 6pt,
    ]{coloredthmversion}

\declaretheoremstyle[
    headformat=\normalfont\textcolor{\thmcolordark}{\bfseries\NAME\,\NUMBER}\NOTE,%
    bodyfont=\normalfont\itshape,
    spaceabove = 6pt,
    spacebelow = 6pt,
    ]{coloredthm}

\declaretheoremstyle[
    headformat=\normalfont\textcolor{\thmcolordark}{\bfseries\NAME\,\NUMBER}\NOTE,%
    bodyfont=\normalfont,
    spaceabove = 6pt,
    spacebelow = 6pt,
    ]{coloreddef}

\iftoggle{customthms}
{
    \theoremstyle{coloredthmversion}
}
{}

\iftoggle{customthms}{
  \theoremstyle{coloredthm}
  \newtheorem{theorem}{Theorem}
  \newtheorem{lemma}{Lemma}[section]
  \newtheorem{corollary}{Corollary}[section]
  \newtheorem{proposition}[lemma]{Proposition}
}
{}

\newtheorem*{thminformal*}{Informal Theorem}

\iftoggle{customthms}{
    \theoremstyle{coloreddef}
    \newtheorem{definition}{Definition}[section]

    \newtheorem{property}{Property}[section]
}
{
  
}

\newtheorem{assumption}{Assumption}[section]
\newtheorem{condition}{Condition}[section]

\makeatletter
\newcommand{\neutralize}[1]{\expandafter\let\csname c@#1\endcsname\count@}
\makeatother

\iftoggle{customthms}{
    \newtheoremstyle{named}{}{}{\itshape}{}{\bfseries}{}{.5em}{\Cref{#3} {\normalfont (informal)} }{}
    \theoremstyle{named}
    
    \theoremstyle{plain}
}
{}

\newtheorem*{theorem*}{Theorem}
\newtheorem*{lemma*}{Lemma}
\newtheorem*{corollary*}{Corollary}
\newtheorem*{proposition*}{Proposition}
\newtheorem*{claim*}{Claim}
\newtheorem*{fact*}{Fact}
\newtheorem*{observation*}{Observation}
\newtheorem*{definition*}{Definition}
\newtheorem*{remark*}{Remark}
\newtheorem*{example*}{Example}

\def\bz{\mathbf{z}}

\def\ddefloop#1{\ifx\ddefloop#1\else\ddef{#1}\expandafter\ddefloop\fi}
\def\ddef#1{\expandafter\def\csname bb#1\endcsname{\ensuremath{\mathbb{#1}}}}
\ddefloop ABCDEFGHIJKLMNOPQRSTUVWXYZ\ddefloop

\def\ddefloop#1{\ifx\ddefloop#1\else\ddef{#1}\expandafter\ddefloop\fi}
\def\ddef#1{\expandafter\def\csname frak#1\endcsname{\ensuremath{\mathfrak{#1}}}}
\ddefloop ABCDEFGHIJKLMNOPQRSTUVWXYZ\ddefloop

\def\ddefloop#1{\ifx\ddefloop#1\else\ddef{#1}\expandafter\ddefloop\fi}
\def\ddef#1{\expandafter\def\csname fr#1\endcsname{\ensuremath{\mathfrak{#1}}}}
\ddefloop ABCDEFGHIJKLMNOPQRSTUVWXYZ\ddefloop

\def\ddefloop#1{\ifx\ddefloop#1\else\ddef{#1}\expandafter\ddefloop\fi}
\def\ddef#1{\expandafter\def\csname eul#1\endcsname{\ensuremath{\EuScript{#1}}}}
\ddefloop ABCDEFGHIJKLMNOPQRSTUVWXYZ\ddefloop

\def\ddefloop#1{\ifx\ddefloop#1\else\ddef{#1}\expandafter\ddefloop\fi}
\def\ddef#1{\expandafter\def\csname scr#1\endcsname{\ensuremath{\mathscr{#1}}}}
\ddefloop ABCDEFGHIJKLMNOPQRSTUVWXYZ\ddefloop

\def\ddefloop#1{\ifx\ddefloop#1\else\ddef{#1}\expandafter\ddefloop\fi}
\def\ddef#1{\expandafter\def\csname b#1\endcsname{\ensuremath{\mathbf{#1}}}}
\ddefloop ABCDEFGHIJKLMNOPQRSTUVWXYZ\ddefloop

\def\ddefloop#1{\ifx\ddefloop#1\else\ddef{#1}\expandafter\ddefloop\fi}
\def\ddef#1{\expandafter\def\csname bhat#1\endcsname{\ensuremath{\hat{\mathbf{#1}}}}}
\ddefloop ABCDEFGHIJKLMNOPQRSTUVWXYZ\ddefloop

\def\ddefloop#1{\ifx\ddefloop#1\else\ddef{#1}\expandafter\ddefloop\fi}
\def\ddef#1{\expandafter\def\csname btil#1\endcsname{\ensuremath{\tilde{\mathbf{#1}}}}}
\ddefloop ABCDEFGHIJKLMNOPQRSTUVWXYZ\ddefloop

\def\ddefloop#1{\ifx\ddefloop#1\else\ddef{#1}\expandafter\ddefloop\fi}
\def\ddef#1{\expandafter\def\csname bst#1\endcsname{\ensuremath{\mathbf{#1}^\star}}}
\ddefloop ABCDEFGHIJKLMNOPQRSTUVWXYZ\ddefloop

\def\ddefloop#1{\ifx\ddefloop#1\else\ddef{#1}\expandafter\ddefloop\fi}
\def\ddef#1{\expandafter\def\csname bst#1\endcsname{\ensuremath{\mathbf{#1}^\star}}}
\ddefloop abcdeghijklmnopqrstuvwxyz\ddefloop

\def\ddefloop#1{\ifx\ddefloop#1\else\ddef{#1}\expandafter\ddefloop\fi}
\def\ddef#1{\expandafter\def\csname bhat#1\endcsname{\ensuremath{\hat{\mathbf{#1}}}}}
\ddefloop abcdefghijklmnopqrstuvwxyz\ddefloop

\def\ddefloop#1{\ifx\ddefloop#1\else\ddef{#1}\expandafter\ddefloop\fi}
\def\ddef#1{\expandafter\def\csname b#1\endcsname{\ensuremath{\mathbf{#1}}}}
\ddefloop abcdeghijklnopqrstuvwxyz\ddefloop

\def\ddefloop#1{\ifx\ddefloop#1\else\ddef{#1}\expandafter\ddefloop\fi}
\def\ddef#1{\expandafter\def\csname barb#1\endcsname{\ensuremath{\bar{\mathbf{#1}}}}}
\ddefloop abcdefghijklmnopqrstuvwxyz\ddefloop

\def\ddef#1{\expandafter\def\csname c#1\endcsname{\ensuremath{\mathcal{#1}}}}
\ddefloop ABCDEFGHIJKLMNOPQRSTUVWXYZ\ddefloop
\def\ddef#1{\expandafter\def\csname h#1\endcsname{\ensuremath{\widehat{#1}}}}
\ddefloop ABCDEFGHIJKLMNOPQRSTUVWXYZ\ddefloop
\def\ddef#1{\expandafter\def\csname hc#1\endcsname{\ensuremath{\widehat{\mathcal{#1}}}}}
\ddefloop ABCDEFGHIJKLMNOPQRSTUVWXYZ\ddefloop
\def\ddef#1{\expandafter\def\csname t#1\endcsname{\ensuremath{\widetilde{#1}}}}
\ddefloop ABCDEFGHIJKLMNOPQRSTUVWXYZ\ddefloop
\def\ddef#1{\expandafter\def\csname tc#1\endcsname{\ensuremath{\widetilde{\mathcal{#1}}}}}
\ddefloop ABCDEFGHIJKLMNOPQRSTUVWXYZ\ddefloop

\usepackage{tcolorbox}
\tcbuselibrary{skins,breakable}

\tcbset{
  aibox/.style={
      width=\linewidth,
      top=6pt,
      bottom=0pt,
      left=1.5pt,
      right=1.5pt,
colback=blue!60!gray!5,
colframe=black,
      colbacktitle=black,
      enhanced,
      center,
      attach boxed title to top left={yshift=-0.1in,xshift=0.15in},
      boxed title style={boxrule=0pt,colframe=white,},
    }
}

\DeclareMathOperator*{\argmin}{arg\,min}

\newcommand{\ballkr}[1][r]{\cB_{k}(r)}

\DeclareMathSymbol{\shortminus}{\mathbin}{AMSa}{"39}

\relax

\usepackage{preamble/color-edits}

\addauthor{ms}{magenta}

\newcommand{\ignore}[1]{}

\iftoggle{arxiv}
{
\input{preamble/arxiv_title}
}
{}

\title{Compression Forcing: Hierarchical History Tokenization Enables Flexible World Models}

\author{\small
Siqiao Huang\footnote{\texttt{\{siqiaoh, parthak, mchen5, hengkaip, kgeng, ochehab, msimchow\}@andrew.cmu.edu} \label{foot:cmu}}\textsuperscript{,}\footnote{\texttt{huang-sq23@mails.tsinghua.edu.cn} \label{foot:thu}}\textsuperscript{,$\$$,$*$} ~
Partha Kaushik\textsuperscript{\ref{foot:cmu},$*$} ~
Michael Chen\textsuperscript{\ref{foot:cmu},$*$} ~
Hengkai Pan\textsuperscript{\ref{foot:cmu},$*$} ~
Kaiwen Geng\textsuperscript{\ref{foot:cmu},$*$} \\
\small
Omar Chehab\textsuperscript{\ref{foot:cmu}} ~
Fernando Moreno-Pino\footnote{\texttt{fernando.moreno-pino@bristol.ac.uk} \label{foot:fernando1}}\textsuperscript{,\textit{\textcolor{darkcontrarycolor}{d}}} ~
Max Simchowitz\textsuperscript{\ref{foot:cmu}} \\
\vspace{-.3em}
\rule{.38\textwidth}{.7pt}
\\
\footnotesize
$^{\$}$Project lead.\quad$^{*}$Equal contribution. \\
\footnotesize $^{a}$Carnegie Mellon University\quad 
$^{b}$Tsinghua University\quad
$^{c}$University of Bristol\quad
$^{d}$University of Oxford ~~
}

\paperdate{\today}

\date{\vspace{-.5em}}

\begin{document}

\begin{tcolorbox}[
colback=blue!60!gray!5, colframe=gray!50, 
boxrule=0pt, 
    arc=2mm
]
\maketitle
\tcbline
\begin{minipage}[t]{1.0\linewidth}
    \centering
\includegraphics[width=\textwidth]{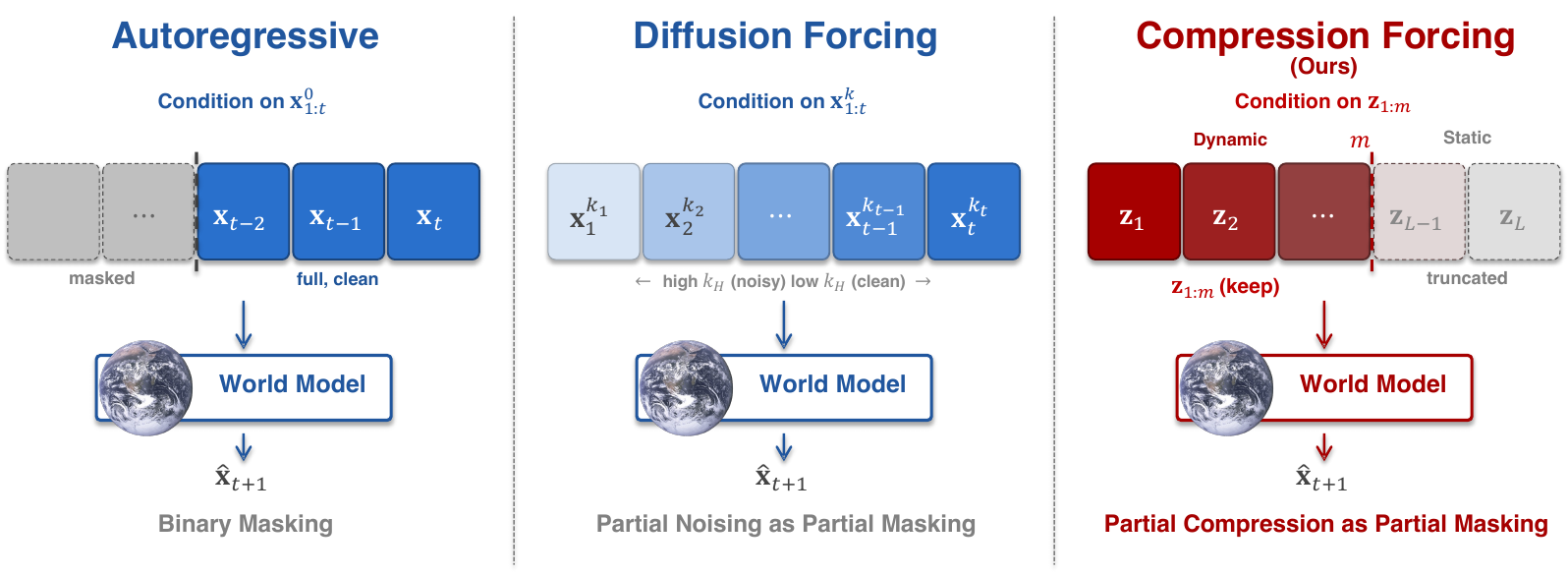}
             \captionof{figure}{
        \small  \textbf{Comparison of history conditioning strategies}. (a) Autoregressive models use \textit{binary masking} of history frames (discarding frames out of context window entirely); (b) Diffusion Forcing utilizes \textit{partial noised} history latents; (c) Compression Forcing introduces \textit{partial compression as partial masking}. By tokenizing history based on dynamic saliency, the world model can flexibly condition on a truncated prefix ($z_{1:m}$), allowing selective amplification of dynamic history information.} 
    \label{fig:single_column_image}
\end{minipage}

  \makeatletter
  \tcbline \ifdefempty{\metadatalist}{}{\metadatalist\par}
  \makeatother
  
\end{tcolorbox}
\begin{abstract}
Effective world modeling requires compressing salient history information that facilitates future prediction. However, most existing approaches inherit tokenization and conditioning mechanisms from video generation, where objectives are centered around perceptual realism rather than predictive utility. This discrepancy limits the world model's effectiveness, especially in environments dominated by visual nuisances or task-irrelevant background details. To this end, we introduce \textit{Compression Forcing} (CF), a general framework that aligns history compression with the predictive objective of world modeling. Compression Forcing features \textit{Dynamics-Aware Tokenization} (\DAT), a novel tokenization method that compresses history observations into a sequence ordered by decreasing dynamics saliency. Built on a diffusion backbone, this structure enables \textit{Hierarchical History Guidance} (\HHG), allowing selective amplification of dynamics-salient information during inference. Extensive experiments across various environments demonstrate improved prediction fidelity, physical realism, and downstream decision-making performance, establishing Compression Forcing as a highly flexible and effective approach for world modeling.
\end{abstract}

\newpage 

\section{Introduction}

World models \citep{Ha2018WorldM, LeCun2022APT} have emerged as a cornerstone of intelligent decision-making, enabling agents to plan and act by simulating future trajectories conditioned on past observations and actions \citep{dreamer, hafner2025training}. Recently, by adopting architectural designs and training paradigms from video generation models, world models have achieved unprecedented generative and predictive capabilities across diverse domains \citep{huang2025vid2world0,guo2025ctrl,pai2025mimicvideovideoactionmodelsgeneralizable,ye2026worldactionmodelszeroshot,kim2026cosmos}. Despite remarkable progress, the objectives of the two paradigms are fundamentally distinct: video generation optimizes for perceptual fidelity and visual realism, whereas world modeling demands faithfully capturing environment dynamics to support accurate downstream prediction and planning. This distinction raises a natural question: \emph{From first principles, how should we design a sequential generative model specifically tailored to the requirements of world model applications?}

The use of historical information (e.g. frames) is a central consideration. Standard autoregressive approaches treat history in a \textit{binary} manner: frames within the context window are fully conditioned upon, while those outside are entirely discarded. Diffusion Forcing \citep{chen2024diffusion, song2025history} introduces a more flexible alternative by enabling conditioning on partially noised histories at inference time. By interpolating between isotropic Gaussian noise and clean observations, this essentially acts as a frequency-based low-pass filter \citep{dieleman2024spectral} of history information, treating \textit{partial noising as partial masking}. While this mechanism aligns well with perceptual objectives in video generation, where low-frequency structure dominates human sensitivity, it is less aligned with the predictive purposes of world modeling. 

Instead, we introduce an alternative principle: \textit{partial compression as partial masking}. Rather than relying on human-imposed prior for masking, we learn an ordered tokenization where hierarchy is induced by a task-specific objective. By conditioning on variable-length prefixes of this learned sequence, truncation naturally serves as a structured form of partial masking, aligning information decomposition in a data-centric fashion. As a concrete realization of this principle in the context of world modeling, we introduce \textcolor{black}{{\textit{Dynamics-Aware Tokenization}}} (\textcolor{primalcolor}{\textbf{\texttt{DAT}}}), a tokenization method compressing historical information into a hierarchical sequence ordered by dynamic saliency. Through training via a task-specific objective, \textcolor{primalcolor}{\textbf{\texttt{DAT}}} prioritizes dynamic motions in earlier tokens, whereas static visual information is encapsulated in later tokens.

\begin{figure*}[t]
    \centering
    \includegraphics[width=\linewidth]{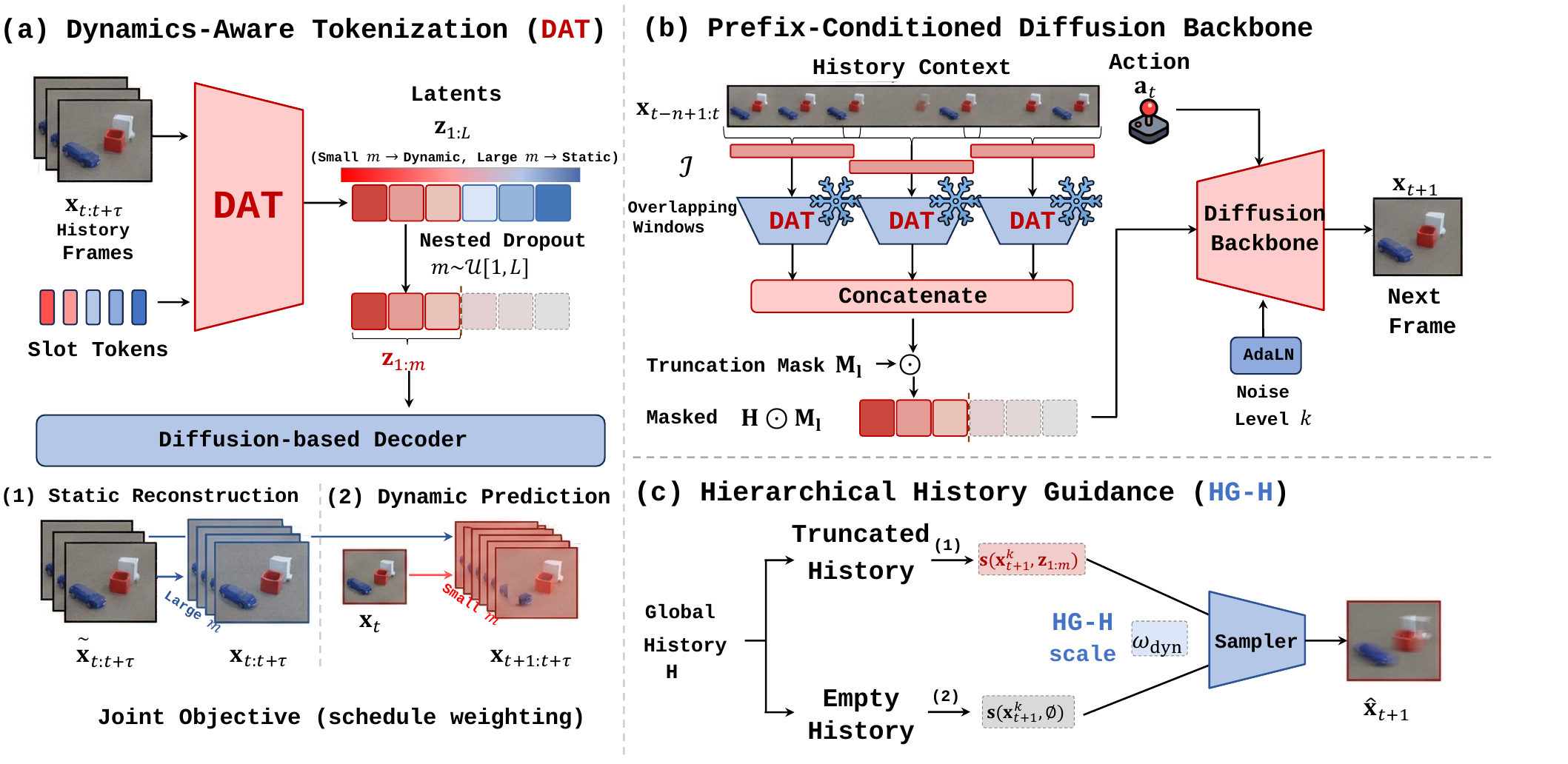}
    \caption{\textbf{Overview of the Compression Forcing Framework}. (a) \DAT\ encodes history context into a hierarchical token sequence, with leading tokens capturing dynamics and trailing tokens capturing static details. (b) The Diffusion Backbone predicts the next frame conditioned on arbitrary-length prefixes of \DAT-encoded latents and the agent's action. (c) \HHG\ composes score estimates from varying-length prefixes, enabling inference-time amplification of dynamically salient history information.}
    \label{fig:cf_arch}
\end{figure*}

Building upon this structured tokenization, we further introduce \textit{Compression Forcing} (CF), a training and sampling framework that operationalizes partial compression as partial masking within diffusion-based world models. Specifically, CF trains a diffusion-based backbone to condition on variable-length prefixes of \DAT\ tokens, enabling prediction under partial history compression (\Cref{fig:single_column_image}), used as a residual stream to make dynamics-relevant information accessible to network attention queries (\Cref{fig:cf_arch}, \Cref{sec:backbone}).  Crucially, this architectural design unlocks a novel inference-time capability: \textit{Hierarchical History Guidance} (\HHG, \Cref{sec:hierarchical_history_guidance}). By composing score functions derived from the full history sequence and \DAT's leading subset, the model is equipped with the inference-time flexibility to selectively amplify dynamic-relevant history, effectively suppressing visual nuisances to ensure robust world modeling.


Through extensive experiments, we demonstrate that Compression Forcing achieves high-fidelity future prediction across a diverse set of environments and action spaces, including dynamic video forecasting \citep{carreira2018short}, 3D game simulation \citep{pearce2022counter}, and real-world robotic manipulation \citep{open_x_embodiment_rt_x_2023, rt_1}. Moreover, beyond perceptual metrics, we show that our world model translates to stronger downstream policy evaluation \citep{li2025evaluating} performance, verifying Compression Forcing's effectiveness in aiding downstream decision-making.

Our main contributions are as follows: (1) We propose \textit{partial compression as partial masking}, a general principle that replaces noise-based history conditioning with learned, task-driven hierarchical compression, applicable beyond world modeling. (2) We instantiate this principle through \textit{Compression Forcing}, a framework comprising \textit{Dynamics-Aware Tokenization} (\DAT) for saliency-ordered history compression, a diffusion-based backbone for prefix-flexible conditioning, and \textit{Hierarchical History Guidance} (\HHG) for inference-time amplification of dynamics-critical information. (3) We validate the effectiveness of our approach across various domains, demonstrating improved prediction fidelity and effective downstream decision-making performance.
\section{Preliminaries and Related Work}
    
    
    

\begin{figure}[t]
    \centering
    \includegraphics[width=\linewidth]{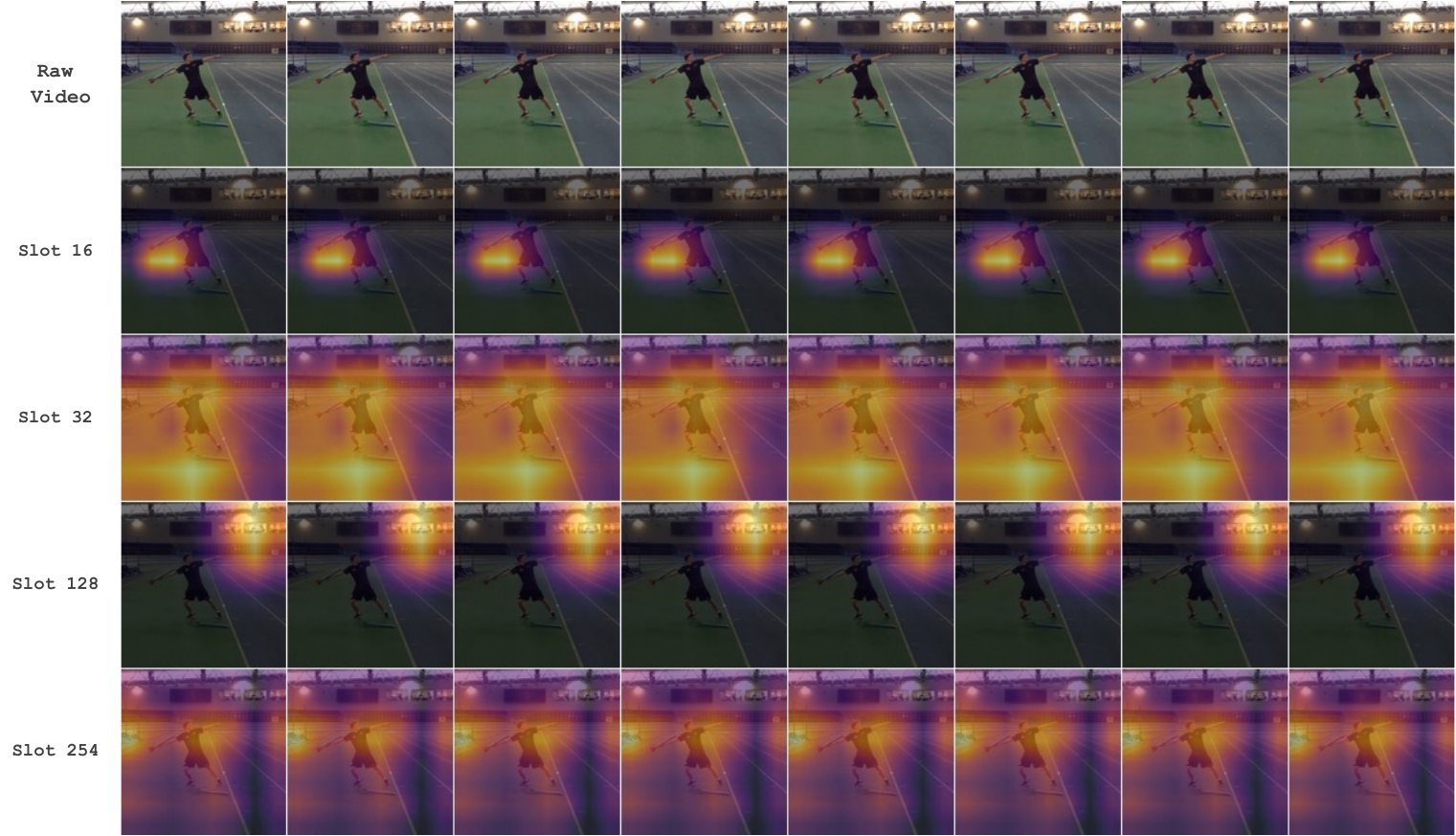}
    \caption{\textbf{Tokenization Attention map}. We visualize attention maps in the cross-attention layer between \DAT\ tokenized tokens and image patches in the dynamic decoder. As the slot index increases, cross-attention shifts from regions with dynamic motion (the athlete in motion) to static objects in the background.}
    \label{fig:attention_map}
\end{figure}

\subsection{World Models}
World Models \citep{LeCun2022APT, Ha2018WorldM} are predictive models learned by agents to capture the dynamics of the environment, commonly formulated as Partially-Observable Markov Decision Processes (POMDPs) \citep{pomdp}. Concretely, a POMDP is specified by the tuple $(\mathcal{S}, \mathcal{O}, \Omega, \mathcal{A}, \mathcal{T}, r, \gamma)$, where the environment evolves over an underlying Markovian state $s_t \in \mathcal{S}$ according to a transition function $\mathcal{T}: \mathcal{S} \times \mathcal{A} \rightarrow \Delta(\mathcal{S})$. Since the states $s_t$ are inaccessible, agents rely on observations $o_t \sim \Omega \left( \cdot \mid s_t\right)$ to perceive the environment. Consequently, the primary objective of world models is to approximate the environment dynamics by estimating the distribution of future observations conditioned on history observations and actions: $ p_\theta (o_{t+1}| o_{\leq t}, a_{\leq t})$. In this paper, we focus on pixel-space observations $\mathcal{O}\equiv \mathbb{R}^{H\times W\times C}$, and denote each frame as $\bx_t$, following conventions in diffusion-based generation.

\subsection{Diffusion Models}
Diffusion Models \citep{ho2020denoising, songdenoising} generate high-fidelity samples by learning to reverse a gradual noise-injection process. Formally, the forward process gradually converts data $\bx^0 \sim q$ into noise $\bx^K \sim \mathcal{N}(\mathbf{0}, \mathbf{I})$ through a Markov Chain, defined as \begin{equation}
    q(\bx^k \mid \bx^{k-1}) = \mathcal{N}\left(\sqrt{1-\beta^k}\bx^{k-1}, \beta_k \mathbf{I}\right),
\end{equation}
where $\beta_{0:K}$ is a pre-defined schedule of noise levels. Sampling is implemented via a reverse process, given by \begin{equation}
    p_\theta \left( \bx^{k-1} \mid \bx^k \right) = \mathcal{N}\left( \boldsymbol{\mu}_\theta( \bx^k, k), \gamma_k \mathbf{I}\right).
\end{equation}
Rather than explicitly modeling the transition densities, modern approaches often reparameterize this by learning to approximate the noise component added at each step $\boldsymbol{\epsilon}^k = (\sqrt{1-\overline{\alpha}_k})^{-1}( \bx^{k}-
\sqrt{\overline{\alpha}_k}\ \bx^{0})$, where $\alpha_k = 1-\beta_k$ and $\overline{\alpha}_k = \prod_{i=1}^k \alpha_i$. The training objective is then simplified to minimizing the mean square loss:\begin{equation}
    \mathcal{L}(\theta)=\mathbb{E}_{k,\boldsymbol{\epsilon}, \bx^0}[||\boldsymbol{\epsilon} -\boldsymbol{\epsilon}_\theta(\bx^k,k)||^2].
\end{equation}
Sampling is performed by iterative ancestral denoising: 
\begin{equation}
    \bx^{k-1}\leftarrow \frac{1}{\sqrt{\alpha_k}}(\bx^k -  \frac{1-\alpha_k}{\sqrt{1-\overline{\alpha}_k}}\boldsymbol{\epsilon}_\theta(\bx^k,k))+\sigma_k \mathbf{w}, \text{where } \mathbf{w}\sim \mathcal{N}(\mathbf{0}, \mathbf{I}).
\end{equation}

\subsection{Sequential Video Generation and History Guidance} 
Sequential video generation predicts future frames $\bx_\mathcal{G}^0$ conditioned on history observations $\bx_\mathcal{H}^0$, enabling long-context video generation with flexible horizon \citep{huang2025selfforcing,cui2025self0forcing000} and on-the-fly streaming \citep{Yin2024FromSB, yang2025longlive}. Beyond standard teacher forcing approaches \citep{yan2021videogpt}, a prevailing paradigm utilizes \textit{Diffusion Forcing} \citep{chen2024diffusion} to enable flexible history conditioning compatible with diffusion architectures. By sampling independent noise levels for each frame during training, it allows for diffusion-based auto-regressive rollouts conditioned on clean history. Building on this, DFoT \citep{song2025history} proposes to treat the history sequence as an external control signal \citep{ho2022classifier0free}, steering future generation by composing score functions conditioned on history interpolated with isotropic Gaussian noise of varying levels $k_\mathcal{H}$:
\begin{equation}
    \nabla \log p_k(\bx_{\mathcal{G}}^k | \bx_{\mathcal{H}}^0) + \omega \big[\nabla \log p_k\big(\bx_{\mathcal{G}}^k | \bx_{\mathcal{H}}^{k_{\mathcal{H}}}\big) - \nabla \log p_k(\bx_{\mathcal{G}}^k)\big].
\end{equation}
By operationalizing guidance through noised history tokens $\bx_\mathcal{H}^{k_\mathcal{H}}$, this history-conditioning mechanism acts as a \textit{frequency-based low-pass filter} to enforce conditioning on low-frequency components in history observations. While this frequency-based prior is well-suited for general video synthesis, it is less suited for world modeling, where context differentiation should be governed by \textit{predictive salience} (i.e., dynamics) rather than \textit{spatial frequency}.

\subsection{Vision Tokenization}
Visual tokenization serves as a foundational component of modern generative architectures, compressing high-dimensional observations into computationally tractable representations for backbone processing. Classical generative approaches, such as SD-VAE \citep{Rombach_2022_CVPR} and VQGAN \citep{esser2020taming}, focus on optimizing perceptual fidelity through continuous regularization or discrete codebooks. Concurrently, emerging research \citep{zheng2025diffusion} demonstrates that representation encoders commonly seen in VLMs \citep{oquab2023dinov2, tschannen2025siglip}, originally designed for semantic understanding, are also highly effective tokenizers for generative tasks. On the structural front, approaches like Semanticist \citep{semanticist} and ElasticTok \citep{yan2024elastictok0} explore hierarchical tokenization via nested dropout, enabling adaptive reconstruction with variable token counts. However, while recent world models have shown impressive results \citep{bar2024navigation, alonso2024diffusion}, they predominantly rely on tokenizers inherited from video generation models, allocating a \textit{fixed, uniform} number of tokens per frame regardless of their predictive importance. The potential of vision tokenization methods specifically tailored for the predictive needs of world models remains largely unexplored.
\section{Compression Forcing}
In this section, we introduce \textit{Compression Forcing}, a general framework for constructing flexible world models via hierarchical history tokenization. Departing from standard approaches \citep{alonso2024diffusion, bar2024navigation, huang2025vid2world0} that employ history tokenization methods which allocates fixed, uniform budgets for each frame, Compression Forcing compresses history into sequences ordered by dynamic saliency, enabling world modeling with flexible amplification of dynamics-salient history information at inference time.

We first introduce \textit{Dynamics-Aware Tokenization} (\DAT), a novel tokenization method that compresses historical observations into hierarchical token sequences ordered by dynamic saliency (Section~\ref{sec:tokenizer}). Building on this structured representation, we detail the design of our \textit{Diffusion-based World Model Backbone}, designed to condition on arbitrary-length prefixes of the \DAT \ compressed token sequences (Section~\ref{sec:backbone}). Finally, we demonstrate that these designs unlock a novel inference-time capability: \textit{Hierarchical History Guidance} (\HHG), a guidance mechanism that allows selective amplification of the role of dynamics-critical history information in generation guidance to facilitate flexible future prediction (Section~\ref{sec:hierarchical_history_guidance}).

\subsection{Dynamics-Aware Tokenizer (\textcolor{primalcolor}{\texttt{DAT}})}
\label{sec:tokenizer}
Standard vision tokenizers in video generation models~\citep{vqvae, esser2020taming, Rombach_2022_CVPR} are optimized for pixel-level reconstruction objectives, resulting in representations that allocate uniform budgets per frame. Especially combined with history guidance methods that interpolate between isotropic Gaussian noise and clean images, this creates a \textit{``frequency-based bias"} \citep{dieleman2024spectral} where guidance is performed by enforcing conditioning on low-frequency components, i.e. treating \textit{partial noising as partial masking}. While this human-imposed prior aligns well with perceptual objectives in video generation, it is misaligned with the predictive demands of world modeling. We propose a more general alternative: \textit{partial compression as partial masking}. Rather than relying on a fixed frequency-based decomposition, we learn a tokenization whose hierarchy is induced by \textit{dynamic saliency}, so that truncating the token sequence naturally serves as a task-aligned form of partial masking that effectively retains dynamics-critical information.

\paragraph{\textbf{Dynamics-Aware Objective.}} To realize this principle, we design \textit{Dynamics-Aware Tokenization} (\DAT), a novel tokenization method that compresses history frames into a sequence ordered by dynamic saliency, disentangling dynamic motions from static visual nuisances. We achieve this through two complementary decoding objectives: static reconstruction and dynamic prediction, whose relative weighting across variable-length prefixes induces the desired hierarchy. Concretely, an encoder $\mathcal{E}$ compresses the history frames $\bx_{t:t+\tau}$ into a latent sequence $\bz_{1:L} = \mathcal{E}(\bx_{t:t+\tau})$,and a diffusion-based decoder $\mathcal{D}$ is trained to decode variable-length prefixes $\bz_{1:m}$ ($m \le L$) under two distinct objectives:
\begin{itemize}
    \item[(a)] \textbf{Static Reconstruction:} Given only the prefix $\bz_{1:m}$ and no external context, the decoder is trained to reconstruct the full clip $\textcolor{lightblue}{\bx_{t:t+\tau}}$. Since the prefix alone must account for all visual content, this compels tokens to capture a lossless compression of information:
    \begin{equation}
        \label{eq:static}
        \textcolor{lightblue}{\mathcal{L}_{\text{static}}} = \mathbb{E}_{k, \epsilon} \Big[ \big\| \textcolor{lightblue}{\bx_{t:t+\tau}} - \mathcal{D}(\bz_{1:m}, \textcolor{lightblue}{\bx^k_{t:t+\tau}}, \ba_{t:t+\tau-1}, k) \big\|^2 \Big].
    \end{equation}
    
    \item[(b)] \textbf{Dynamic Prediction:} Given the prefix $\bz_{1:m}$ and the initial frame $\bx_t$ as additional context, the decoder is trained to predict target frames $\textcolor{primalcolor}{\bx_{t+1:t+\tau}}$ given $\bz_{1:m}$. Essentially, $\bx_t$ allows the model to act as a residual learner, enabling the prefix to only focus on dynamic changes:
    \begin{equation}
        \label{eq:dyn}
        \textcolor{primalcolor}{\mathcal{L}_{\text{dyn}}} = \mathbb{E}_{k, \epsilon} \Big[ \big\| \textcolor{primalcolor}{\bx_{t+1:t+\tau}} - \mathcal{D}(\bz_{1:m}, \textcolor{primalcolor}{\bx^k_{t+1:t+\tau}}, \textcolor{hyppurple}{\bx_t},\ba_{t:t+\tau-1}, k) \big\|^2 \Big].
    \end{equation}
\end{itemize}

We enforce the desired hierarchy through a schedule-weighted combination of these two objectives. During training, we randomly sample prefix lengths $m \in [1, L]$ and minimize:
\begin{equation}
\mathcal{L}_{\text{total}} = \mathbb{E}_{m} \Big[ \omega(m)\ \textcolor{lightblue}{\mathcal{L}_{\text{static}}} + \left(1-\omega(m)\right)\ \textcolor{primalcolor}{\mathcal{L}_{\text{dyn}}} \Big],
\end{equation}
where $\omega(m)$ is a monotonic increasing function. When $m$ is small, the objective is dominated by dynamic prediction $\textcolor{lightblue}{\mathcal{L}_{\text{dyn}}}$, compelling leading tokens to prioritize motion dynamics; as $m$ grows, $\textcolor{primalcolor}{\mathcal{L}_{\text{static}}}$ takes over, shifting focus toward full visual reconstruction. This yields a naturally ordered sequence: earlier tokens encode \textit{dynamic changes}, while later tokens progressively capture \textit{static visual details}.

\paragraph{\textbf{Tokenizer Architecture.}}
The tokenizer $\mathcal{E}$ is built based on the Video Vision Transformer (ViViT) \citep{arnab2021vivit} architecture. To produce the ordered sequence $\bz_{1:L}$, we append $L$ learnable \textit{Slot Tokens} at the end of the patchified video sequence.  Information flow is regulated through a \textit{Hybrid Attention Mechanism}: video patch tokens attend bidirectionally to each other but are masked from attending to slot tokens, whereas the $i$-th slot token $\bz_i$ attends globally to all video patches but \textit{causally} to preceding slots $\bz_{<i}$. This causal design is essential for ensuring that prefix-truncation is well-structured: since $\bz_i$ is computed without access to succeeding slots $\bz_{>i}$, any prefix $\bz_{1 :m}$ remains a self-consistent representation regardless of the truncation point.

For the diffusion decoder $\mathcal{D}$, we adopt a video diffusion transformer architecture \citep{ma2024latte}. Notably, rather than maintaining separate networks for each objective, we implement the Static and Dynamic Decoders as a \textit{single model} with shared parameters. The distinction between tasks is achieved through noise conditioning \citep{chen2024diffusion}: for Static Reconstruction, we apply identical noise levels to all frames; for Dynamic Prediction, we set the noise level of the condition frame $\bx_t$ to zero and identical noise levels to other frames.

\newcommand{\Henc}{\mathbf{H}^{\mathrm{enc}}}
\subsection{Prefix-Conditioned World Model Backbone}
\label{sec:backbone}
\paragraph{\textbf{Backbone Architecture.}}
Building upon the hierarchical history representation produced by \DAT, we instantiate the world model backbone $\mathcal{W}_\theta$ as a conditional diffusion transformer \citep{peebles2023scalable, bar2024navigation}. The backbone predicts the next frame $\bx_{t+1}$ conditioned on the agent's recent actions $\ba_{{t}}$ and observations $\bx_{t-n+1:t}$. Within each transformer block, we employ a dual-conditioning mechanism: the fine-grained history representation $\mathbf{H}$ is injected via cross-attention. Meanwhile, the low-dimensional conditions (action $\ba_t$ and diffusion noise level $k$) are projected into a joint embedding $\xi = \text{MLP}_\text{act}(\ba_t) + \text{MLP}_\text{noise}(k)$, modulating intermediate features $\mathbf{h}$ via AdaLN \citep{xu2019understanding}:
\begin{equation}
    \text{AdaLN}(\mathbf{h}, \xi) = \gamma(\xi) \odot \text{LayerNorm}(\mathbf{h}) + \beta(\xi),
\end{equation}
where $\gamma(\cdot)$ and $\beta(\cdot)$ are learned scale and shift parameters regressed from $\xi$.

\paragraph{\textbf{History Prefix Conditioning.}} Since \DAT\ accepts a fixed-length input of $\tau$ frames, a discrepancy exists between its limited receptive field and the backbone's need for broader history context (where $n > \tau$). To bridge this gap, we encode the history context $\bx_{t-n+1:t}$ using a \textit{sliding window approach} with a temporal stride $s$. We extract $\mathcal{I} = \lfloor \frac{n-\tau}{s} \rfloor + 1$ overlapping frame sequences. Each window $\bx^{(i)}$is independently tokenized by the frozen \DAT\ encoder, and the resulting latents are concatenated to form an \emph{encoded} global history context $\Henc$:
\begin{equation}
\Henc = \text{Concat}\left( \mathcal{E}(\bx^{(1)}), \mathcal{E}(\bx^{(2)}), \dots, \mathcal{E}(\bx^{(\mathcal{I})}) \right) \in \mathbb{R}^{(\mathcal{I} \cdot L) \times D}.
\end{equation}

To fully leverage the saliency ordering within $\mathbf{H}$, the world model backbone must learn to condition on variable-length prefixes of each history window. During training, for each encoded window $\mathcal{E}(\bx^{(i)})$, we independently sample a  \textbf{truncation budget} $l_t \sim \mathcal{U}[0, L]$ and construct a binary mask $\mathbf{M}_{l_i}$ that preserves the leading $l_i$ tokens and masks out the rest. The training objective is defined as:
\begin{equation}
    \mathcal{L}_{\text{backbone}} = \mathbb{E}_{k, \epsilon, \mathbf{l}} \left[ \| \epsilon - \epsilon_\theta(\bx_{t+1}^k, k, \Henc \odot \mathbf{M}_\mathbf{l}, \ba_t) \|^2 \right],
\end{equation}
where $\mathbf{l}=[l_1,...,l_\mathcal{I}]$ is the \textbf{budget vector} and $\mathbf{M}_\mathbf{l}=\text{Concat}(\mathbf{M}_{l_1},...,\mathbf{M}_{l_\mathcal{I}})$. This training objective equips the backbone with the ability to condition on any prefix length of the the \DAT-encoded history, laying the foundation for controllable guidance of history information at inference time detailed in Section \ref{sec:hierarchical_history_guidance}.
\begin{figure*}[t]
    \centering
    \includegraphics[width=\linewidth]{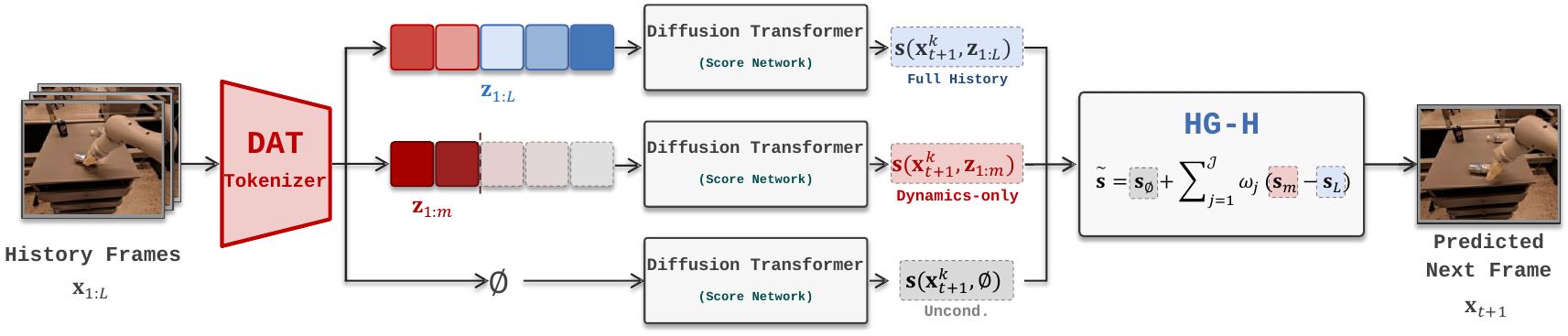}
    \caption{\textbf{Sampling with Hierarchical History Guidance (\HHG)}. By composing score functions conditioned on the full history representation and its high-saliency leading subset, the model selectively amplifies essential motion dynamics while suppressing visual nuisances, enabling robust prediction.}
    \label{fig:hhg}
\end{figure*}
\subsection{Hierarchical History Guidance (\HHG)}
\label{sec:hierarchical_history_guidance}
Given the world model backbone's flexibility to condition on arbitrary-length prefixes of the \DAT-encoded history, we introduce \textit{Hierarchical History Guidance} (\HHG), an inference-time guidance mechanism that amplifies dynamics-salient history information through score composition~\citep{ho2022classifier0free}.

\paragraph{\textbf{General Guidance Formulation.}} Let $\mathbf{H}^{(\mathbf{l})}$ denote the encoded history context masked by a budget vector $\mathbf{l}$. The diffusion backbone functions as a learned score estimator $\mathbf{s}_\theta(\mathbf{x}_{t+1}, \mathbf{H}^{(\mathbf{l})}, \ba_t, k)\approx\nabla \log p_k(\mathbf{x}_{t+1} | \mathbf{H}^{(\mathbf{l})}, \ba_t)$. Given a set of budget vectors $\{\mathbf{l}_j\}^{\mathcal{J}}_{j=1}$, we can guide generation via composing $\tilde{\nabla} \log p_k(\mathbf{x}_{t+1})$ as a weighted sum of conditional scores:
\begin{align*}
    \tilde{\nabla} \log p_k(\bx_{t+1}^k) =& \nabla \log p_k(\bx_{t+1}^k | \varnothing, \ba_t) \\
    &+ \sum_{j=1}^\mathcal{J} \omega_j \cdot \left[ \nabla \log p_k(\bx_{t+1}^k | \mathbf{H}^{(\mathbf{l}_j)}, \ba_t) - \nabla \log p_k(\bx_{t+1}^k | \varnothing, \ba_t) \right],
\end{align*}
where $\omega_j$ controls the guidance strength for each component. Essentially, this formulation allows \textit{flexible amplification} on specific historical information components \textit{based on their predictive utility}.

\paragraph{\textbf{Hierarchical History Guidance.}} As the primary instantiation of this guidance formulation, we propose \textit{Hierarchical History Guidance}(\HHG), prioritizing high-saliency dynamics over visual details. For every $j\in\{1,...,\mathcal{J}\}$, we assign a \textit{uniform saliency budget} $\mathbf{l}_{j}^\text{dyn} = [m_j, \dots, m_j]$, where $m_j \leq L$ represents the truncation threshold across all history windows. The guidance objective simplifies to:
\begin{align*}
\tilde{\nabla} \log p_k(\bx_{t+1}^k) =& \nabla \log p_k(\bx_{t+1}^k | \varnothing, \ba_{t})
    \\ &+ \sum_{j=1}^{\mathcal{J}-1} \omega_j \cdot \left[ \nabla \log p_k(\bx_{t+1}^k | \mathbf{H}^{\mathbf{l}_{j+1}^\text{dyn}}, \ba_{t}) - \nabla \log p_k(\bx_{t+1}^k |\mathbf{H}^{\mathbf{l}_{j}^\text{dyn}}, \ba_{t}) \right].
\end{align*}
By guiding generation with truncated prefixes of each history window, \HHG\ amplifies the influence of motion dynamics encoded in the leading tokens, while suppressing stochastic visual nuances captured by later tokens. Essentially, this operationalizes the \textit{partial compression as partial masking} principle: the saliency ordering imposed by \DAT\ ensures that prefix truncation serves as a task-aligned form of partial masking, with $\omega_j$s controlling the degree of amplification at inference time.

\section{Experiments}

In this section, we evaluate Compression Forcing under three distinct world modeling environments, and compare their performance against state-of-the-art baselines. We demonstrate competitive performance against existing methods in a variety of tasks (Section \ref{subsec:kinetics}, \ref{subsec:csgo} and \ref{subsec:rt1}) and conduct in-depth analysis on tokenization and generation error distribution  to understand the effects of \DAT\ and \HHG\ (Section \ref{subsec:analysis}). Experimental Details can be found in Appendix \ref{app:exp_details}.

\subsection{Experimental Setting}
\paragraph{\textbf{Environments.}} We evaluate Compression Forcing across three domains: general video prediction, game simulation, and robot manipulation. For video prediction, we use Kinetics-600 \citep{carreira2018short}, a large-scale benchmark with diverse, motion-rich videos, utilizing it for history-conditioned video prediction. For game simulation, we adopt \textit{Counter-Strike: Global Offensive} (CS:GO), using a 5.5M-frame online gameplay dataset \citep{pearce2022counter}. For robot manipulation, we use RT-1 \citep{open_x_embodiment_rt_x_2023}, comprising real-world manipulation tasks such as picking and placing. The action spaces differ across domains: Kinetics-600 is action-free; CS:GO involves discretized keyboard and mouse inputs; and RT-1 uses a 7D continuous end-effector action space. Together, these environments span a broad spectrum of dynamics complexity, action dependence, and visual diversity, enabling a comprehensive evaluation of Compression Forcing across both virtual and real-world settings. Further details can be found in Appendix \ref{app:env_details}.

\paragraph{\textbf{Evaluation Metrics.}} 
We evaluate video prediction performance using a variety of widely-adopted metrics, measuring pixel-level fidelity and perceptual similarity. For pixel-level fidelity, we utilize PSNR \citep{hore2010image}, and for perceptual similarity, we report SSIM\citep{wang2004image}, FVD \citep{unterthiner2018towards}, FID \citep{heusel2017gans} and LPIPS \citep{zhang2018unreasonable}. Details are shown in Appendix \ref{app:metrics_details}. Finally, we evaluate fidelity to physical dynamics through two downstream metrics deferred to Appendix~\ref{app:ext_exp}: (1) object tracking fidelity, where we sample query points within SAM-generated object masks, track them with CoTracker3 across ground-truth and predicted videos, and measure point-trajectory deviations and penalties for lost tracks, and (2) policy evaluation correlation via Real2Sim evaluation protocol of \citet{li2025evaluating}, capturing how reliable Compression Forcing is in predicting and ranking policy performance.


\paragraph{\textbf{Baselines.}} We compare Compression Forcing against two distinct sets of baselines. The first shares our backbone architecture but conditions on alternative latent spaces for history compression: SD-VAE \citep{Rombach_2022_CVPR}, which compresses each frame into a fix-budget representation, and ElasticTok \citep{yan2024elastictok0}, which compresses video segments into a 1D sequence with leading tokens prioritizing reconstruction fidelity, akin to a learned PCA. The second comprises strong existing world models, including DIAMOND \citep{alonso2024diffusion} in CS:GO and DreamerV4 \citep{hafner2025training} in RT-1. Baseline details are shown in Appendix \ref{app:baseline_details}.

\subsection{Compression Forcing for Dynamic Video Generation}
\label{subsec:kinetics}

\begin{table}[t]
\centering
\caption{\textbf{Dynamic Video Prediction on Kinetics-600.} We evaluate against baselines sharing the diffusion-based world model backbone architecture and training budget; rows differ only in history tokenization method (SD-VAE, ElasticTok, or \DAT) and whether history guidance is applied. Details can be found in Appendix \ref{app:kinetics_details}.}
\label{tab:kinetics}
\setlength{\tabcolsep}{7pt}
\begin{tabular}{ll ccccc}
\toprule
Tokenizer & Guidance & SSIM$\uparrow$ & PSNR$\uparrow$ & FID$\downarrow$ & LPIPS$\downarrow$ \\
\midrule
SD-VAE & \ding{55} & 0.543 & 18.53 & 90.57 & 0.351 \\
ElasticTok & \ding{55} & 0.544 & 18.55 & 91.23 & 0.351 \\
ElasticTok & \ding{51} & 0.546 & 18.60 & 90.74 & 0.349 \\
\DAT & \ding{55} & 0.551 & 18.77 & 89.26 & 0.344 \\
\midrule
\rowcolor{gray!10} \textbf{\DAT} & \textbf{\HHG} & \textbf{0.553} & \textbf{18.83} & 
\textbf{88.13} & \textbf{0.343} & \\
\bottomrule
\end{tabular}
\end{table}

We first evaluate Compression Forcing on action-free video prediction using Kinetics-600 \citep{carreira2018short}, a large-scale benchmark featuring diverse, motion-rich natural videos. This setting is particularly informative as it isolates our \DAT\ from action conditioning: the hierarchical structure must emerge purely from visual dynamics, instead of relying on actions to aid dynamic prediction. \textcolor{black}{Across all baselines, we train the diffusion-based backbone for 160k gradient steps. We train the \DAT\ tokenizer for 120k gradient steps and the ElasticTok tokenizer for 175k gradient steps. During inference, we condition our model on 16 history frames, and autoregressively rollout 8 frames.}  Further training and inference details are provided in Appendix~\ref{app:kinetics_details}.

As shown in Table \ref{tab:kinetics}, Compression Forcing consistently outperforms all existing baselines in all metrics evaluated. Even without guidance, where the model is exposed to all \DAT\ tokens, the performance of our model consistently outperforms SD-VAE and ElasticTok, showcasing the efficacy of leveraging dynamics-aware tokenization for precise prediction. Performance gains yielded by \HHG\ further demonstrate that partial compression serves as an effective mechanism for targeted probability steering during inference.



\begin{figure*}[t]
    \centering
    \includegraphics[width=\linewidth]{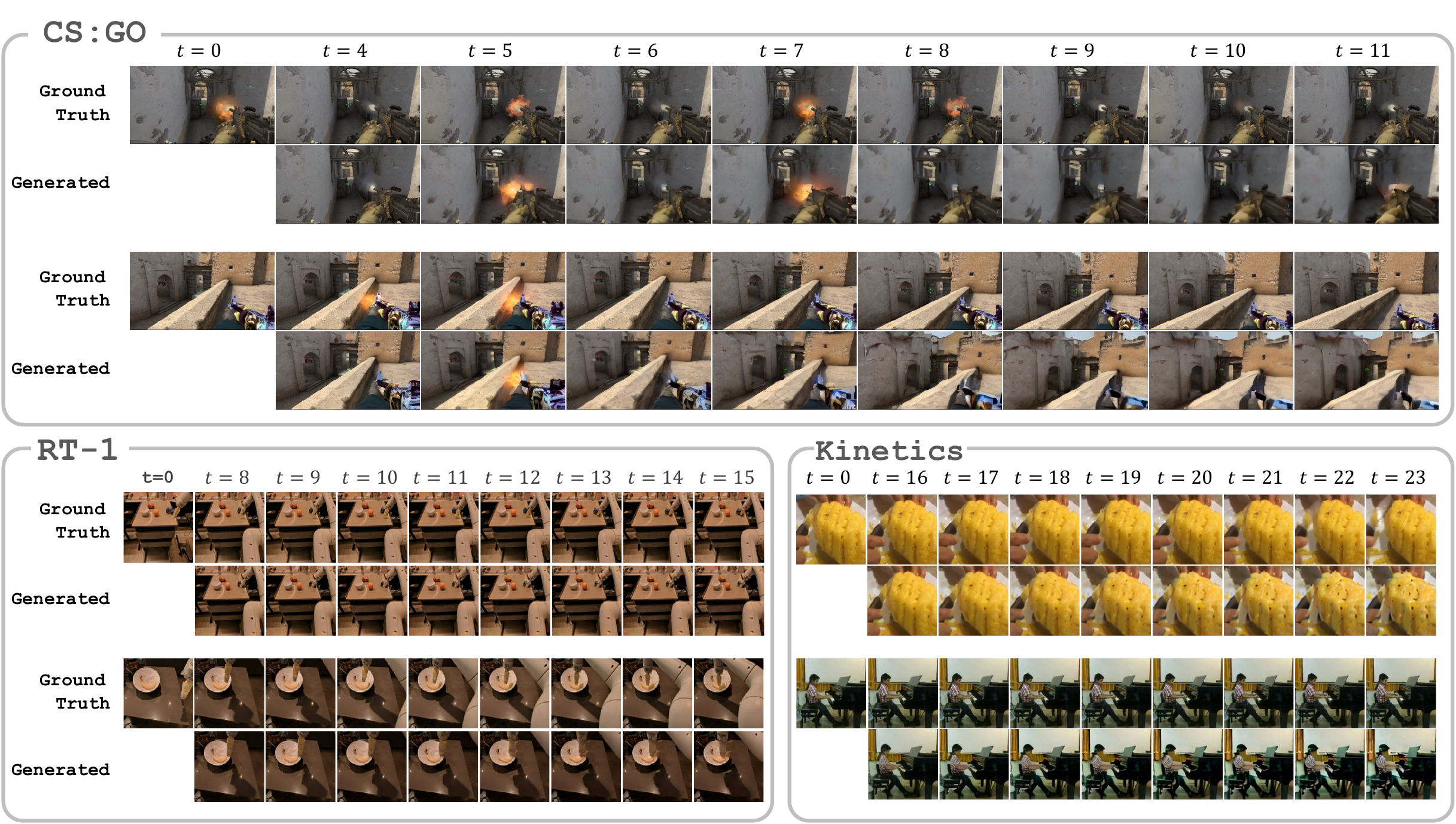}
    \caption{\textbf{Qualitative Generation Results on Multiple Domains.} Compression Forcing enables accurate and high-fidelity auto-regressive predictions in CS:GO, RT-1 and Kinetics. Zoom in for details. Extended visualizations in Appendix \ref{app:ext_exp}.}
    \label{fig:vis}
\end{figure*}

\subsection{Compression Forcing for Game Simulation}
\label{subsec:csgo}

\begin{table}[t]
\centering
\caption{\textbf{Game Simulation on CS:GO.} We evaluate against DIAMOND \citep{alonso2024diffusion}, a strong established baseline, and same-architecture variants differing only in history tokenization (SD-VAE, ElasticTok). For all methods, we auto-regressively roll out 12 frames from 4 context frames. Details in Appendix \ref{app:csgo_details}.}
\setlength{\tabcolsep}{8pt}
\label{table:csgo}
\begin{tabular}{l ccccc}
\toprule
Method & SSIM$\uparrow$ & PSNR$\uparrow$ & FVD$\downarrow$ & FID$\downarrow$ & LPIPS$\downarrow$ \\
\midrule
DIAMOND-Fast & 0.449 & 18.2 & 577.1 & 115.6 & 0.547 \\
DIAMOND-HQ & 0.447 & \textbf{18.3} & 368.5 & 87.2 & 0.510 \\
SD-VAE & \textbf{0.496} & 17.29 & 231.9 & 70.25 & 0.514 \\
ElasticTok(\HHG) & 0.494 & 17.22  & 197.1 & 74.2 & 0.519 \\
\midrule
\rowcolor{gray!10} \DAT(\HHG) & 0.479 & 17.24 & \textbf{123.0} & \textbf{53.6} & \textbf{0.493}\\
\bottomrule
\end{tabular}
\end{table}

We next evaluate Compression Forcing on action-conditioned game simulation using \textit{Counter-Strike: Global Offensive} (CS:GO), a popular 3D first-person shooting game. This environment is particularly challenging due to its highly stochastic dynamics and partial observability, including rapid viewpoint changes, occlusions, and complex player-environment interactions. We adopt the online gameplay dataset of \citet{pearce2022counter} (5.5M frames, 95 hours) and compare against DIAMOND~\citep{alonso2024diffusion}, a strong autoregressive diffusion-based world model. All other baselines share the same backbone architecture and training budget of 550k gradient steps. Both tokenizers ElasticTok and \DAT\ were trained for 100k gradient steps, for the same token dimensions and count per window. During inference, we condition on 4 context frames and autoregressively roll out 12 predicted frames. We use \HHG\ for models with ElasticTok and \DAT\ tokenizers. Further details are provided in Appendix~\ref{app:csgo_details}. 

As shown in Table \ref{table:csgo}, \DAT\ outperforms the existing popular baseline DIAMOND and other tokenization methods using the same diffusion-backbone architecture and training budget. The results for models with ElasticTok and \DAT\ showcase the effectiveness of \HHG, and its amplification when combined with \DAT. 

\begin{table}[t]
\centering
\caption{\textbf{Robot Manipulation on RT-1.} We evaluate against baselines sharing the diffusion-based world model backbone architecture and training budget (except DIAMOND and Dreamer 4); rows differ only in history tokenization method (SD-VAE, ElasticTok, or \DAT) For all methods, we auto-regressively roll out 16 frames from 8 context frames. Details in Appendix \ref{app:rt1_details}.}
\setlength{\tabcolsep}{8pt}
\label{table:rt1}
\begin{tabular}{l ccccc}
\toprule
Method & SSIM$\uparrow$ & PSNR$\uparrow$ & FVD$\downarrow$ & FID$\downarrow$ & LPIPS$\downarrow$ \\
\midrule
SD-VAE & \textbf{0.722} & \textbf{20.47} & 120.4 & 42.82 & 0.267 \\
ElasticTok(\HHG) & 0.721 & 20.42  & 105.4 & 44.86 & 0.267 \\
DIAMOND & 0.6130 & 14.05 & 382.4 & 104.50 & 0.3267 \\
Dreamer 4 & 0.2718 & 16.27 & 984.1 & 365.67 & 0.6050 \\
\midrule
\rowcolor{gray!10} \textbf{\DAT}(\HHG) & 0.721 & 20.43 & \textbf{101.5} & \textbf{42.77} & \textbf{0.266}\\
\bottomrule
\end{tabular}
\end{table}

\begin{figure*}[t]
    \centering
    \includegraphics[width=\linewidth]{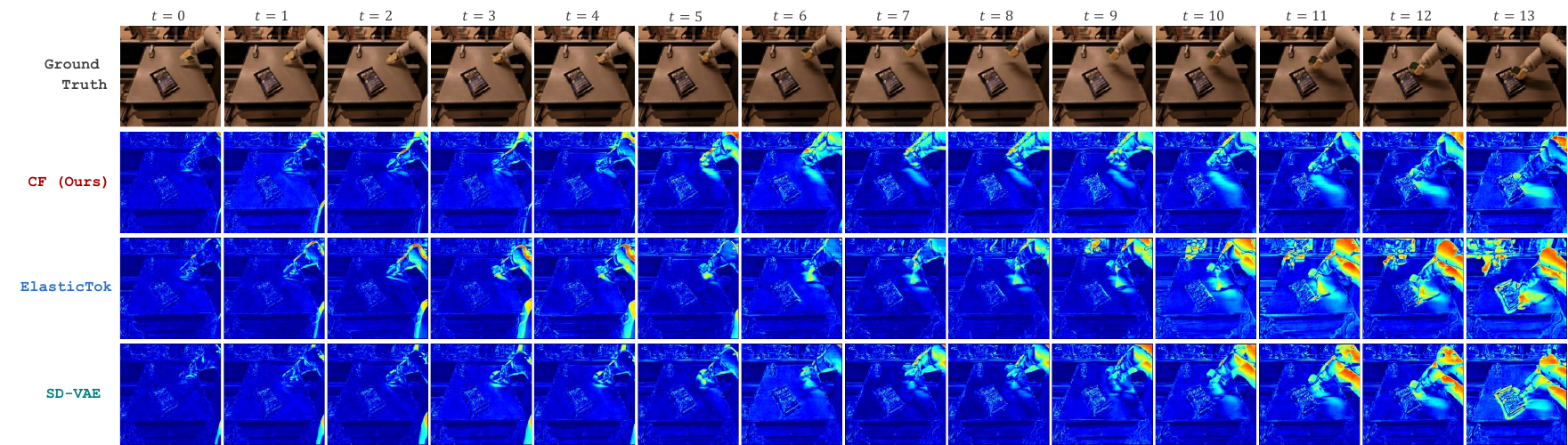}
    \caption{\textbf{Spatial Distribution of Prediction Errors.} While ElasticTok and SD-VAE accumulate error around the robot end-effector and contact points, regions most critical for accurate physical interaction, Compression Forcing maintains high fidelity in these dynamic regions across the full prediction horizon.}
    \label{fig:error_analysis}
\end{figure*}

\subsection{Compression Forcing for Robot Manipulation}
\label{subsec:rt1}

Robot Manipulation is another great task to measure a world model's capability, especially in real environments and physical constraints provided by the RT-1 dataset \citep{open_x_embodiment_rt_x_2023}. The RT-1 dataset contains 130k episodes and 700+ distinct tasks. We use the same setup for baselines as in CS:GO and train \DAT\ and the other baselines for 80k gradient steps. Both tokenizers ElasticTok and \DAT\ were trained for 40k gradient step, with same number of tokens and their dimension. For evaluation we roll out 16 generated frames from a context of 8 frames. We compare our method against two world-modeling paradigms: DIAMOND \citep{alonso2024diffusion} a diffusion-based world model and Dreamer 4 \citep{hafner2025training}, a latent dynamics RL agent. Further details and extended visualizations are provided in Appendix~\ref{app:rt1_details} and Appendix \ref{sec:full_rt1_vis}.

As shown in Table \ref{table:rt1}, we observe a substantial performance gap between these established baselines and our diffusion-backbone variants. While DIAMOND and Dreamer 4 struggle with structural drift and temporal blurring in the RT-1 domain, Compression Forcing maintains high-fidelity reconstructions and motion continuity.



\subsection{Analysis}
\label{subsec:analysis}

\paragraph{\textbf{Emergence of Hierarchical Features in Tokenization.}} To verify whether our Dynamics-Aware Tokenization (\DAT)\ learns hierarchical representations ordered by dynamic saliency, we visualize cross-attention maps between \DAT\ tokens and image patches in the dynamic decoder. As shown in Figure~\ref{fig:attention_map}, the attention maps exhibits a clear saliency-ordered hierarchy. Leading tokens assign high attention weights to regions of active motion (the athlete in motion), while higher-index slots progressively shift toward static background elements such as the court boundaries. This confirms that our joint schedule-weighted objective induces \DAT\ to decouple dynamic interactions from visual nuisances, effectively encoding historical information into hierarchical representations.

\paragraph{\textbf{Spatial Distribution of Prediction Errors.}} To assess whether this hierarchical tokenization translates to improved prediction in dynamics-critical regions, we analyze per-pixel prediction error across autoregressive rollouts on RT-1 (Figure~\ref{fig:error_analysis}). While baseline tokenizers (SD-VAE and ElasticTok) accumulate substantial and progressive errors around the manipulated target object and the robot's end-effector, regions most critical for accurate physical interaction, Compression Forcing maintains markedly low error in these dynamic regions throughout the prediction horizon, showcasing practical benefits of aligning history compression with predictive saliency rather than perceptual fidelity. Additional visualizations and further analysis can be found in Appendix \ref{sec:full_analysis}



\section{Conclusion}

We introduce \textit{Compression Forcing}, a novel world modeling framework for built upon the principle of \textit{partial compression as partial masking}. Through the co-design of Dynamics-Aware Tokenization (\DAT), a prefix-conditioned diffusion backbone, and Hierarchical History Guidance (\HHG), our approach compresses history into sequences ordered by dynamic saliency and selectively amplifies dynamics-critical information at inference time. Extensive experiments across real-world video forecasting, 3D game simulation, and robotic manipulation demonstrate that Compression Forcing achieves improved prediction fidelity and downstream decision-making performance compared to prior methods, showcasing the broad applicability of our method. 

While we believe this work marks significant progress in rethinking world model design principles in a task-centric manner, there is substantial room for improvement. Promising directions include extending prediction beyond single next-frame to multi-step \citep{bar2024navigation} or variable-horizon \citep{huang2025vid2world0,huang2026nano} futures, and differentiating truncation budgets between recent and distant history to better capture their distinct roles in prediction.  More broadly, we hope that the principle of partial compression as partial masking serves as a useful foundation for future exploration in world models and beyond.

\newpage
\newpage
\bibliographystyle{plainnat}
\bibliography{refs}
\newpage 

\newpage
\appendix

\tableofcontents
\newpage

\section{Theoretical Analysis}
\label{app:theory}

We provide a linear motivation for \DAT's residual formulation by comparing low-rank projections of dynamics residuals and raw future sequences.

\paragraph{Setup.}
Let $\bx_t \in \mathbb{R}^d$ denote the current observation and $\mathbf{Y} \triangleq \bx_{t+1:t+\tau} - \bx_t \in \mathbb{R}^{d \times \tau}$ the ground-truth dynamics residual (stacked as columns). We model the raw future as $\mathbf{F} \triangleq \bx_{t+1:t+\tau} = \bx_t \mathbf{1}^\top + \mathbf{Y}$, i.e.\ a static component plus a dynamics component. We assume access to $N$ i.i.d.\ samples $\{(\bx_t^{(n)}, \mathbf{F}^{(n)}, \mathbf{Y}^{(n)})\}_{n=1}^N$, and define the empirical uncentered second-moment matrices:
\begin{align}
    \mathbf{C}_F &\triangleq \frac{1}{N}\sum_{n=1}^N \text{vec}(\mathbf{F}^{(n)}) \, \text{vec}(\mathbf{F}^{(n)})^\top, \label{eq:cov_f} \\
    \mathbf{C}_Y &\triangleq \frac{1}{N}\sum_{n=1}^N \text{vec}(\mathbf{Y}^{(n)}) \, \text{vec}(\mathbf{Y}^{(n)})^\top, \label{eq:cov_y}
\end{align}
where $\text{vec}(\cdot)$ denotes column-wise vectorization. We then denote the static uncentered second-moment matrix as $\mathbf{C}_S \triangleq \frac{1}{N}\sum_n \text{vec}(\bx_t^{(n)} \mathbf{1}^\top)\,\text{vec}(\bx_t^{(n)} \mathbf{1}^\top)^\top$. 

\begin{assumption}[Static-Dynamics Decomposition]\label{assump:decomp}
For future sequence decomposition of $\mathbf{F} = \mathbf{S} + \mathbf{Y}$, where $\mathbf{S} = \bx_t \mathbf{1}^\top$ is the static component and $\mathbf{Y} = \bx_{t+1:t+\tau} - \bx_t$ is the dynamics residual, we assume that the cross second moment $\frac{1}{N}\sum_n \mathrm{vec}(\mathbf{S}^{(n)})\,\mathrm{vec}(\mathbf{Y}^{(n)})^\top$ vanishes, which implies $\mathbf{C}_F = \mathbf{C}_S + \mathbf{C}_Y$.
\end{assumption}

\paragraph{Two projection problems.}
In the projection objectives and errors below, $\mathbf{F}$ and $\mathbf{Y}$ denote their vectorized forms, and expectations are taken over the empirical sample distribution.
We consider learning a rank-$m$ orthogonal projector $\mathbf{P} \in \mathbb{R}^{d\tau \times d\tau}$ (with $\mathbf{P}^\top = \mathbf{P}$, $\mathbf{P}^2 = \mathbf{P}$, $\text{rank}(\mathbf{P}) = m$) that best preserves the data under ridge regularization:
\begin{itemize}
    \item \textbf{Raw projection} (analogous to ElasticTok / standard tokenization):
    \begin{equation}
        \mathbf{P}^*_F = \argmin_{\text{rank}(\mathbf{P})=m} \; \mathbb{E}\big[\|\mathbf{F} - \mathbf{P}\,\mathbf{F}\|^2\big] + \lambda \|\mathbf{P}\|_F^2.
        \label{eq:raw_proj}
    \end{equation}
    \item \textbf{Residual projection} (analogous to \DAT's dynamic prediction objective):
    \begin{equation}
        \mathbf{P}^*_Y = \argmin_{\text{rank}(\mathbf{P})=m} \; \mathbb{E}\big[\|\mathbf{Y} - \mathbf{P}\,\mathbf{Y}\|^2\big] + \lambda \|\mathbf{P}\|_F^2.
        \label{eq:res_proj}
    \end{equation}
\end{itemize}

\begin{proposition}[Dynamics prediction gap]\label{prop:gap}
Under Assumption~\ref{assump:decomp}, assume $\mathbf{C}_S \neq \mathbf{0}$, $\lambda > 0$, $1 \le m < d\tau$, and $\lambda_m(\mathbf{C}_Y) > \lambda_{m+1}(\mathbf{C}_Y)$, where eigenvalues are ordered in descending order. The dynamics prediction error satisfies:
\begin{equation}
    \mathbb{E}\big[\|\mathbf{Y} - \mathbf{P}^*_Y \mathbf{Y}\|^2\big] \;\leq\; \mathbb{E}\big[\|\mathbf{Y} - \mathbf{P}^*_F \mathbf{Y}\|^2\big],
    \label{eq:gap}
\end{equation}
with equality if and only if the selected top-$m$ eigenspaces of $\mathbf{C}_F$ and $\mathbf{C}_Y$ coincide (i.e.\ the static component introduces no spectral interference).
\end{proposition}

\noindent To formally prove Proposition~\ref{prop:gap}, we first establish two supporting lemmas regarding the properties of orthogonal projectors and second-moment matrices.

\begin{lemma}[Projection Error as Trace]\label{lem:trace}
For any random vector $\mathbf{z}$ with finite second moment $\mathbf{C} = \mathbb{E}[\mathbf{z}\mathbf{z}^\top]$, the expected reconstruction error using an orthogonal projection matrix $\mathbf{P}$ is given by:

\begin{equation}
    \mathbb{E}\big[\|\mathbf{z} - \mathbf{P}\mathbf{z}\|^2\big] = \text{tr}\big((\mathbf{I} - \mathbf{P})\mathbf{C}\big).
\end{equation}
\end{lemma}
\begin{proof}
Using the property $\|\mathbf{v}\|^2 = \text{tr}(\mathbf{v}\mathbf{v}^\top)$ and the idempotence of the orthogonal projector $(\mathbf{I} - \mathbf{P})^\top(\mathbf{I} - \mathbf{P}) = (\mathbf{I} - \mathbf{P})$, we have:
\begin{align}
    \mathbb{E}\big[\|\mathbf{z} - \mathbf{P}\mathbf{z}\|^2\big] &= \mathbb{E}\big[\text{tr}\big((\mathbf{z} - \mathbf{P}\mathbf{z})(\mathbf{z} - \mathbf{P}\mathbf{z})^\top\big)\big] \nonumber \\
    &= \mathbb{E}\big[\text{tr}\big((\mathbf{I} - \mathbf{P})\mathbf{z}\mathbf{z}^\top(\mathbf{I} - \mathbf{P})\big)\big] \nonumber \\
    &= \text{tr}\big((\mathbf{I} - \mathbf{P})\mathbb{E}[\mathbf{z}\mathbf{z}^\top]\big) = \text{tr}\big((\mathbf{I} - \mathbf{P})\mathbf{C}\big).
\end{align}
\end{proof}

\begin{lemma}[Optimality of Eigenspace]\label{lem:pca}
For the regularized rank-$m$ projection problem:
\begin{equation}
    \argmin_{\text{rank}(\mathbf{P})=m} \mathbb{E}[\|\mathbf{z} - \mathbf{P}\mathbf{z}\|^2] + \lambda \|\mathbf{P}\|_F^2,
\end{equation}
the optimal solution is $\mathbf{P}^* = \mathbf{U}_m \mathbf{U}_m^\top$, where $\mathbf{U}_m$ spans the top-$m$ eigenvectors of the uncentered second-moment matrix $\mathbf{C}$.
\end{lemma}
\begin{proof}
For any rank-$m$ orthogonal projector, the Frobenius norm squared is constant: \begin{equation}
    \|\mathbf{P}\|_F^2 = \text{tr}(\mathbf{P}^\top\mathbf{P}) = \text{tr}(\mathbf{P}) = m.
\end{equation}
Thus, the regularization term $\lambda m$ does not affect the optimization landscape. Then by Lemma~\ref{lem:trace}, the problem reduces to:
\begin{equation}
    \argmin_{\text{rank}(\mathbf{P})=m}\text{tr}((\mathbf{I} - \mathbf{P})\mathbf{C}).
\end{equation}
By the Eckart-Young-Mirsky theorem, this trace is minimized when $\mathbf{P}$ projects onto the subspace spanned by the eigenvectors corresponding to the $m$ largest eigenvalues of $\mathbf{C}$.
\end{proof}

Empowered with these Lemmas, we then complete the proof of Proposition~\ref{prop:gap}.

\begin{proof}[Proof of Proposition~\ref{prop:gap}]
Let $\{(\sigma_i^Y, \mathbf{u}_i^Y)\}_{i=1}^{d\tau}$ denote the eigendecomposition of $\mathbf{C}_Y$, with eigenvalues sorted in descending order. 

By Lemma~\ref{lem:pca}, the residual projector $\mathbf{P}^*_Y$ is the global minimizer for the reconstruction error of $\mathbf{Y}$ among all rank-$m$ projectors. Its column space $\mathbf{U}_m^Y$ exactly spans the top-$m$ eigenvectors of $\mathbf{C}_Y$. Applying Lemma~\ref{lem:trace}, the expected error achieves the theoretical minimum:
\begin{equation}
    \mathbb{E}\big[\|\mathbf{Y} - \mathbf{P}^*_Y \mathbf{Y}\|^2\big] = \mathrm{tr}\big((\mathbf{I} - \mathbf{P}^*_Y)\mathbf{C}_Y\big) = \sum_{i=m+1}^{d\tau} \sigma_i^Y.
    \label{eq:err_min}
\end{equation}

Now consider the raw projector $\mathbf{P}^*_F$, which projects onto $\mathbf{U}_m^F$ (the top-$m$ eigenspace of $\mathbf{C}_F$). Using Lemma~\ref{lem:trace}, its expected reconstruction error on the dynamics residual $\mathbf{Y}$ is:
\begin{equation}
    \mathbb{E}\big[\|\mathbf{Y} - \mathbf{P}^*_F \mathbf{Y}\|^2\big] = \mathrm{tr}\big((\mathbf{I} - \mathbf{P}^*_F)\mathbf{C}_Y\big) = \sum_{i=1}^{d\tau} \sigma_i^Y \big(1 - \|(\mathbf{U}_m^F)^\top \mathbf{u}_i^Y\|^2\big).
    \label{eq:err_raw}
\end{equation}

Since $\mathbf{P}^*_Y$ is the global minimizer of $\mathrm{tr}\big((\mathbf{I} - \mathbf{P})\mathbf{C}_Y\big)$ over all rank-$m$ projectors, any other rank-$m$ projector, including $\mathbf{P}^*_F$, must yield error no less than $\mathbf{P}^*_Y$, establishing the inequality:
\begin{equation}
    \mathbb{E}\big[\|\mathbf{Y} - \mathbf{P}^*_Y \mathbf{Y}\|^2\big] \;\leq\; \mathbb{E}\big[\|\mathbf{Y} - \mathbf{P}^*_F \mathbf{Y}\|^2\big].
\end{equation}
The weak inequality follows directly from residual-projector optimality and does not require Assumption~\ref{assump:decomp}; the decomposition only provides an interpretation of the difference between the two objectives.

Under the eigengap assumption, the residual-optimal projector is unique. Equality holds if and only if $\mathrm{span}(\mathbf{U}_m^F) = \mathrm{span}(\mathbf{U}_m^Y)$, which requires $\mathrm{span}(\mathbf{U}_m^Y)$ to be simultaneously an invariant subspace of both $\mathbf{C}_Y$ and $\mathbf{C}_S$.
\end{proof}

\paragraph{Connection to \DAT.}
Proposition~\ref{prop:gap} provides a linear illustration of the intuition behind \DAT's dual-objective design (Section~\ref{sec:tokenizer}). The static reconstruction loss $\mathcal{L}_\text{static}$ (Eq.~\ref{eq:static}) is analogous to the raw projection objective, whose optimal projector $\mathbf{P}^*_F$ is aligned with the full second-moment matrix $\mathbf{C}_F$. The dynamic prediction loss $\mathcal{L}_\text{dyn}$ (Eq.~\ref{eq:dyn}), which conditions on $\bx_t$ and supports residual learning, is analogous to the residual projection objective with optimal projector $\mathbf{P}^*_Y$. By weighting $\mathcal{L}_\text{dyn}$ more heavily for small prefix lengths $m$ (via the schedule $\omega(m)$), \DAT\ encourages the leading tokens to prioritize residual changes. As the prefix grows and $\omega(m)$ increases, $\mathcal{L}_\text{static}$ receives greater weight, encouraging later tokens to recover visual details needed for full reconstruction.

\section{Extended Discussions}

\subsection{Partial Compression as Partial Masking}

While in this paper, we mainly instantiate the principle of \emph{partial compression as partial masking} for world modeling, the applicability of this approach is more general, offering an alternative to existing mechanisms for controlling the information content of conditioning signals in generative models.
 
\paragraph{General formulation.}
Consider a conditional generative model $p_\theta(\bx \mid \mathbf{c})$, where $\mathbf{c}$ denotes an arbitrary conditioning signal. Suppose we have access to an ordered encoder $\mathcal{E}$ that compresses $\mathbf{c}$ into a hierarchical latent sequence $\bz_{1:L} = \mathcal{E}(\mathbf{c})$, where the hierarchy is induced by a task-specific objective such that truncating to a prefix $\bz_{1:m}$ retains the information most relevant to the downstream task. Then, by training $p_\theta(\bx \mid \bz_{1:m})$ to condition on variable-length prefixes $\bz_{1:m}$ for $m \in [0, L]$, truncation naturally serves as a structured form of partial masking. Essentially, the flexibility of conditioning on arbitrary-lengthed prefix, the model is capable of generation with \emph{task-aligned} partial information rather than information filtered by a fixed, human-imposed prior (e.g., frequency-based decomposition in Diffusion Forcing~\cite{chen2024diffusion,dieleman2024spectral}).
 
This formulation immediately enables score-based guidance in the diffusion setting.
Treating $p_\theta$ as a learned score estimator, one can compose score functions across different prefix lengths to selectively amplify specific information components:
\begin{align*}
    \tilde{\nabla} \log p_k(\bx^k) &= \nabla \log p_k(\bx^k \mid \varnothing) \\
    &+ \sum_{j} \omega_j \left[ \nabla \log p_k(\bx^k \mid \bz_{1:m_{j+1}}) - \nabla \log p_k(\bx^k \mid \bz_{1:m_j}) \right].
\end{align*}
 
\paragraph{Task-specific tokenization, task-agnostic guidance.}
A key property of this formulation is the clean separation between two design choices.
The \emph{tokenizer} $\mathcal{E}$ is necessarily task-specific: the ordering it learns must reflect the information hierarchy relevant to the downstream objective. In world modeling, \DAT\ orders tokens by dynamic saliency (Section \ref{sec:tokenizer}); in other domains, the appropriate hierarchy would differ, for instance, key frame is visual retrieval tasks, function groups in protein folding or high-entropy decision-making moments in planning.

In contrast, the \emph{guidance mechanism} above is entirely task-agnostic: once a hierarchical tokenization exists, score composition over variable-length prefixes provides a universal mechanism for inference-time amplification, regardless of what the hierarchy encodes. This separation means that the Compression Forcing framework can serve as inspiration for tasks not limited to world models, where prefix-conditioned generation with hierarchical tokenization can facilitate flexible amplification of control signals for generation.

\subsection{Hierarchical History Guidance}

Notably, our framework can also be integrated with history guidance methods along the time axis \citep{song2025history} by setting different budget $l_i$ for each history segment, which we leave for future work.
The general guidance formulation in Eq.~(11) subsumes several existing conditioning strategies as special cases, and its relationship to prior work merits clarification.
 
\paragraph{Connection to DFoT history guidance.}
DFoT \citep{song2025history} steers future generation by composing score functions conditioned on history observations interpolated with isotropic Gaussian noise of varying levels $k_{\mathcal{H}}$. While in the paper, we mainly discuss the \textit{fractional history guidance}, which can be viewed as a special case of our formulation with 
\begin{equation}
    \mathcal{E}(\mathbf{x}_{t:t+\tau}) = [I_{m/L} (\mathbf{x}_{t:t+\tau})]_{m=0}^L,
\end{equation}
where $I_{k} (\mathbf{x}_{t:t+\tau}) \triangleq (1-k)\cdot\mathbf{x}_{t:t+\tau} + k \cdot\mathbf{w}, \mathbf{w}\sim \mathcal{N}(\mathbf{0}, \mathbf{I})$, as in flow-matching and stochastic interpolants literature \citep{lipman2022, albergo2023building, albergo2023stochastic}.

\paragraph{Composability along the temporal axis.}
While for the sake of simplicity we focus on hierarchical history guidance with uniform budgets for each history window, i.e. $\mathbf{l}_j^{\text{dyn}} = [m_j, \ldots, m_j]$ across all $\mathcal{I}$ windows, our framework naturally extends to heterogeneous truncation budgets across history windows. This is analogous to \textit{temporal history guidance} as in \citet{song2025history}. We leave systematic exploration of such temporally heterogeneous guidance strategies to future work.
 
\paragraph{Connection to classifier-free guidance.}
Classic Classifier-free guidance (CFG)~\cite{ho2022classifier0free}, in for instance text-conditioned image generation, amplifies the influence of a conditioning signal $\mathbf{c}$ by extrapolating between the conditional and unconditional score estimates: $\tilde{\nabla} = (1+\omega)\nabla\log p(\mathbf{x}|\mathbf{c}) - \omega\,\nabla\log p(\mathbf{x})$. This can be viewed as applying a binary mask of the control signal. \HHG\ generalizes this by decomposing $\mathbf{c}$ itself into a hierarchy of sub-conditions $\{\mathbf{H}^{(\mathbf{l}_j)}\}$ and amplifying specific components. When $\mathcal{J}=2$ with $\mathbf{l}_1 = \mathbf{0}$ (empty history) and $\mathbf{l}_2 = [L,\ldots,L]$ (full history), \HHG\ reduces to standard CFG. Building upon hierarchical structures tokenized via \DAT\ , Compression Forcing enables finer-grained control that is not possible for methods treating the conditioning signal as monolithic.

\subsection{Tokenization in World Models}

While most recent pixel-space world models \citep{guo2025ctrl,team2026advancing,huang2025vid2world0} inherit visual tokenization from image and video generation models\citep{svd, wan2025}, earlier work in latent-state and discrete-token-based world modeling has explored representation designs with dynamic awareness. Following the line of works with RSSM-style backbones \citep{dreamer, hafner2024masteringdiversedomainsworld}, Iso-Dream \citep{pan2022isodreamisolatingleveragingnoncontrollable} proposes three separate branches during tokenization, explicitly disentangling controllable, noncontrollable,
and static components from visual data. Following discrete token representation methods in world modeling \citep{vqvae, iris2023}, $\Delta$-IRIS \citep{micheli2024efficient} shows improvements by conditioning the tokenizer on previous frames and actions, intuitively encoding only stochastic deltas between timesteps. However, to the best of our knowledge, \DAT{} is the first to learn a \emph{hierarchical ordering} over tokens by dynamic saliency, rather than a binary separation into predefined categories. This ordering ensures that any truncation point produces a valid representation that smoothly interpolates between dynamics-only and full-detail conditioning, directly enabling variable-rate compression and the inference-time guidance mechanism (\HHG{}) introduced in Section \ref{sec:hierarchical_history_guidance}.

\section{Experiment Details}
\label{app:exp_details}
\subsection{Environment Details}

\textbf{Kinetics-600} \citep{carreira2018short} is a well-known video generation benchmark with $\sim$400k human action videos labeled into 600 classes. We use 128x128 pixel resolution, following other works with this dataset. This dataset is difficult because of the significant diversity among videos. We use it to test the efficacy of our \DAT, backbone, and \HHG\ in an action-free setting. \\\\
\textbf{CS:GO} \citep{pearce2022counter} is a dataset of gameplay videos from the popular video game CS:GO, with actions represented per-timestep as a list of discrete values. We use the largest dataset subset, \texttt{hdf5\_dm\_july2021}, with about 95 hours of diverse online human gameplay on the \textit{Dust II} map.
\\\\
\textbf{RT-1} \citep{rt_1} is a widely used real-world robot experience dataset, with multiple manipulation tasks executed by an embodied robot arm. Episodes are sampled at 3 frames per second. 
\label{app:env_details}
\subsection{Metrics Details}
\label{app:metrics_details}

\textbf{Peak Signal-to-Noise Ratio (PSNR)}: PSNR \citep{hore2010image} measures pixel-level fidelity by comparing the maximum possible pixel intensity to the MSE between a generated and target frame. With $L$ being the maximum possible pixel value, PSNR is calculated with $$\mathrm{PSNR}(x,y)=10\log_{10} \frac{L^2}{\mathrm{MSE}(x,y)}.$$
\textbf{Structural Similarity Index Measure (SSIM)}: SSIM \citep{wang2004image} measures perceived visual qualities. It jointly compares luminance, contrast, and structural similarity rather than raw pixel error alone. For generated and target patches $x$ and $y$, SSIM is defined as $$\mathrm{SSIM}(x,y)=\frac{(2\mu_x\mu_y+C_1)(2\sigma_{xy}+C_2)}{(\mu_x^2+\mu_y^2+C_1)(\sigma_x^2+\sigma_y^2+C_2)}.$$
Here, $\mu_x$ and $\mu_y$ denote the mean intensities of the two patches, $\mu_x^2$ and $\mu_y^2$ are their variances, and $\sigma_xy$ is their covariance. $L$ is dynamic range of the pixel values, and $C_1=(K_1L)^2$ and $C_2=(K_2L)^2$ are  constants for numerical stability. SSIM is particularly useful when two images are similar at a high level but differ in local contrast or structural details, with higher SSIM values indicating greater similarity. Following \citep{huang2025vid2world0}, we compute SSIM over sliding 11x11 Gaussian windows and average those local values into an image-level SSIM. \\\\
\textbf{Learned Perceptual Image Patch Similarity (LPIPS)}: LPIPS \citep{zhang2018unreasonable} measures similarity in a learned feature space. Instead of comparing images through raw intensity differences, it compares deep feature representations extracted from a pretrained network, making it more sensitive to perceptually meaningful differences in texture, structure, and appearance. For generated and target images $x$ and $x'$, LPIPS is computed as
$$
\mathrm{LPIPS}(x,x')=\sum_l \frac{1}{H_lW_l}\sum_{h,w}\left\|w_l \odot \left(\hat{y}^{\,l}_{hw}-\hat{y}^{\,l\prime}_{hw}\right)\right\|_2^2,
$$
where $\hat{y}^{\,l}, \hat{y}^{\,l\prime} \in \mathbb{R}^{H_l \times W_l \times C_l}$ are the channel-wise normalized feature maps of $x$ and $x'$ at layer $l$, $H_l$, $W_l$, and $C_l$ are the spatial height, spatial width, and number of channels of that layer, and $w_l \in \mathbb{R}^{C_l}$ is a learned channel-wise weighting vector. The metric averages squared feature differences across spatial locations and sums them across layers. Lower LPIPS values indicate greater perceptual similarity. \\\\
\textbf{Fr\'echet Inception Distance (FID)}: FID \citep{heusel2017gans} is a distribution-level metric for evaluating the realism of generated images. Rather than comparing each generated sample to a specific target image, it compares the feature distributions of real and generated images after embedding them with a pretrained Inception network. Let $(\mu_r,\Sigma_r)$ and $(\mu_g,\Sigma_g)$ denote the empirical mean and covariance of the real and generated feature embeddings, respectively. FID is defined as
$$
\mathrm{FID}(r,g)=\|\mu_r-\mu_g\|_2^2 + \mathrm{Tr}\!\left(\Sigma_r+\Sigma_g-2(\Sigma_r\Sigma_g)^{1/2}\right).
$$
Here, $\mu_r$ and $\mu_g$ are the means of the real and generated feature distributions, $\Sigma_r$ and $\Sigma_g$ are their covariance matrices, and $\mathrm{Tr}(\cdot)$ denotes the matrix trace. FID is useful because it measures similarity at the distribution level, capturing both sample quality and diversity rather than only pixel-wise agreement. Lower FID values indicate that the generated samples more closely match the real data distribution. \\\\
\textbf{Fr\'echet Video Distance (FVD)}: FVD \citep{unterthiner2018towards} extends the idea of FID from images to videos by comparing distributions of real and generated videos in a learned spatiotemporal feature space. Instead of evaluating frames independently, it embeds entire video clips using a pretrained Inflated 3D ConveNet (I3D) video recognition network, allowing it to account for temporal coherence and motion in addition to visual quality. FVD is formulated identically to FID except that it uses I3D activations instead of Inception activations.

\subsection{Baselines Details}
\label{app:baseline_details}
We compare against two groups of baselines. The first group is designed to isolate the effect of the history representation while keeping the world-model backbone as close as possible to ours. Concretely, these models use the same conditional diffusion backbone, action conditioning, and auto-regressive rollout protocol as Compression Forcing, and differ only in how the history observations are compressed before being provided to the backbone. The second group consists of strong external world-model baselines, namely DIAMOND and DreamerV4, which represent established alternative paradigms for action-conditioned future prediction.\\\\
\textbf{SD-VAE}: This baseline uses the Stable Diffusion VAE \citep{Rombach_2022_CVPR} as the history encoder. Each context frame is encoded independently into a spatial latent map with a fixed per-frame budget, and the diffusion backbone conditions on the stack of these latent tensors directly. As a result, every frame contributes the same amount of unmasked conditioning information regardless of how informative it is for prediction, and the representation is optimized for perceptual reconstruction rather than video-predictive utility.\\\\
\textbf{ElasticTok}: ElasticTok \citep{yan2024elastictok0} is a variable-rate tokenizer that maps visual inputs to an ordered one-dimensional token sequence. In contrast to SD-VAE, it allows the history to be represented by a prefix of varying length, where earlier tokens are intended to carry the most important reconstruction information. In our experiments, we use a frozen pretrained ElasticTok encoder and feed its output tokens to the same diffusion world-model backbone used by our method. Depending on the environment, the history is encoded either frame-wise or over short sliding windows, after which the resulting token sequence is consumed by the backbone through the same cross-attention interface as our tokenizer-based models. ElasticTok thus provides an adaptive and prefix-truncatable representation, but its token ordering is learned from reconstruction objectives, not from the future-prediction objective that drives \DAT. Hierarchical guidance is applied to ElasticTok in the same way as \DAT.\\\\
\textbf{DIAMOND}: DIAMOND \citep{alonso2024diffusion} is an autoregressive diffusion world model that predicts future observations conditioned on recent observations and actions. We use the official public implementation\footnote{\url{https://github.com/eloialonso/diamond}} for this baseline. Unlike \DAT, DIAMOND does not expose a learned hierarchical history representation that can be truncated or selectively amplified at inference time.\\\\
\textbf{DreamerV4}: DreamerV4 \citep{hafner2025training} learns a world model consisting of a causal tokenizer and an action-conditioned dynamics model. Future trajectories are generated by rolling the latent dynamics forward under the action sequence and decoding predicted observations from the latent representations. Since the official implementation is not open-sourced, we use a popular and well-tested community-contributed DreamerV4 implementation\footnote{\url{https://github.com/nicklashansen/dreamer4}}. DreamerV4 is included because it is a strong recent world-model baseline, but it differs substantially from our approach in both training objective and architecture; its latent dynamics model compresses each frame into a flat discrete token sequence with no explicit ordering by predictive importance, while Compression Forcing's \DAT \ learns to order tokens by dynamic saliency.

\subsection{Shared Tokenizer and Backbone configurations.}

The following hyperparameters are shared across the Kinetics, CS:GO, and RT-1 datasets.

\begin{table}[!htbp]
\centering
\caption{\textbf{Shared \DAT\ hyperparameters.} We share our encoder, decoder, diffusion, and training configurations are across environments, as listed in this table, unless stated otherwise in environment-specific sections.}
\label{tab:shared_hyperparams}
\small
\renewcommand{\arraystretch}{1.15}
\begin{tabular}{@{}p{0.55\linewidth}c@{}}
\toprule
\textbf{Hyperparameter} & \textbf{Value} \\
\midrule
\rowcolor{gray!15}
\multicolumn{2}{c}{\textit{Encoder Architecture}} \\
\addlinespace[2pt]
\quad Model & ViViT \\
\quad Depth & 12 \\
\quad Hidden size & 768 \\
\quad Patch size & 16 \\
\quad Attention heads & 12 \\
\quad Tokens per window & 256 \\
\quad Token dimension & 16 \\
\addlinespace[4pt]
\rowcolor{gray!15}
\multicolumn{2}{c}{\textit{Decoder Architecture}} \\
\addlinespace[2pt]
\quad Model & Latte \\
\quad Depth & 12 \\
\quad Hidden size & 384 \\
\quad Patch size & 2 \\
\quad Attention heads & 6 \\
\addlinespace[4pt]
\rowcolor{gray!15}
\multicolumn{2}{c}{\textit{Diffusion}} \\
\addlinespace[2pt]
\quad Timesteps & 1000 \\
\quad Noise schedule & Linear \\
\quad $\beta_1$ & 1e-4 \\
\quad $\beta_T$ & 0.02 \\
\quad Prediction target & $\epsilon$ \\
\quad Latent diffusion & Yes \\
\quad Latent spatial downsample & $8\times$ \\
\quad Latent channels & 4 \\
\quad Latent scaling factor & 0.18215 \\
\addlinespace[4pt]
\rowcolor{gray!15}
\multicolumn{2}{c}{\textit{Training}} \\
\addlinespace[2pt]
\quad Optimizer & AdamW \\
\quad Learning rate & 1e-4 \\
\quad Weight decay & 0 \\
\quad Gradient clip max norm & 0.1 \\
\quad Gradient clip start iteration & 20\,000 \\
\quad GPUs & 8 \\
\quad Float precision & FP32 \\
\addlinespace[4pt]
\rowcolor{gray!15}
\multicolumn{2}{c}{\textit{Data Processing}} \\
\addlinespace[2pt]
\quad Normalization & $[-1, 1]$ \\
\quad Frame stride & 1 \\
\bottomrule
\end{tabular}
\end{table}

\noindent Note that the \DAT \ has an input-independent null token prepended to 255 data-dependent tokens, bringing the total number of tokens to 256. When making unconditional generations from the \DAT \ decoder, we only condition on the null token. The DAT decoder is implemented using $\epsilon$-prediction in the VAE latent space; the $x$-prediction form in Eqs.~(3.1)--(3.2) denotes clean-latent prediction with the corresponding timestep-dependent weighting left implicit, since the two squared-error objectives are equivalent under the standard output reparameterization and SNR-dependent reweighting \citep{kingma2023understanding}.

\paragraph{Shared ElasticTok tokenizer configuration.}
Across all datasets, the ElasticTok tokenizer used hidden size 1024, feed-forward size 2048, 10 encoder layers, 10 decoder layers, 8 attention heads, and RoPE $\theta=5\times 10^6$. Frame windows were encoded into 256 tokens of dimension 16. We trained in FP32 using AdamW with learning rate $1\times 10^{-4}$, weight decay $1\times 10^{-4}$, and 2000 learning-rate warmup steps, with KL weight $1\times 10^{-8}$ and LPIPS loss ratio 0.1. We trained each ElasticTok tokenizer on 8 GPUs.

\paragraph{Shared CDiT backbone configuration.}
Across all datasets, the SDVAE, ElasticTok, and \DAT-conditioned backbones used 12 layers, patch size 2, MLP ratio 4.0, and 4 latent input channels. Diffusion used 1000 timesteps with a linear noise schedule, $\beta_1=1\times 10^{-4}$, $\beta_T=2\times 10^{-2}$, $\epsilon$-prediction, and learned variance. Backbone training used AdamW with learning rate $8\times 10^{-5}$, weight decay 0, and gradient clipping with max norm 10.0. We trained each backbone on a single GPU. 

\subsection{Details of Compression Forcing for Dynamics Video Generation}
\label{app:kinetics_details}

\paragraph{Training.} We trained the ElasticTok and \DAT\ tokenizers to tokenize 8 frame windows. We trained both models to convergence, so we stopped \DAT\ after 120k steps and ElasticTok after 175k steps. The backbones had hidden size 768 and 12 attention heads, and we trained all backbones for 160k steps.

\paragraph{Evaluation.} We condition on 16 context frames and autoregressively roll out 8 future frames. All metrics are computed over 256 evaluation videos. For both our \DAT (\HHG) and ElasticTok with guidance, we use the general guidance formulation with $\vec\omega=(-0.5,1.5)$ and $\vec l=(128,256)$.

\subsection{Details of Compression Forcing for Game Simulation}
\label{app:csgo_details}

\paragraph{Training.}

We trained the ElasticTok and \DAT\ tokenizers to tokenize 4 frame windows. We trained both for 100k steps with batch size 128. The backbones had hidden size 1024 and 16 attention heads, and we trained the backbones with SDVAE, ElasticTok and \DAT\ history conditioning for 550k steps each. 

\paragraph{Evaluation.}

We condition on 4 context frames and autoregressively roll out 12 future frames. All metrics are computed over 500 evaluation videos. For both our \DAT(\HHG) and ElasticTok(\HHG) evaluations, we used the Hierarchical History Guidance formulation with $\vec\omega=(2,4,6)$ and $\vec l = (1,16,128,256)$. 

\subsection{Details of Compression Forcing for Robot Manipulation}
\label{app:rt1_details}

\paragraph{Training.}

We trained the ElasticTok and \DAT\ tokenizers to tokenize 4 frame windows. We trained both for 40k steps with batch size 512. The backbones had hidden size 768 and 12 attention heads, and we trained the backbones with SDVAE, ElasticTok, and \DAT\ history conditioning for 80k steps each.

\paragraph{Evaluation.}

We condition on 8 context frames and autoregressively roll out 16 future frames. All metrics are computed over 500 evaluation videos. For both our \DAT(\HHG) and ElasticTok(\HHG) evaluations, we used the Hierarchical History Guidance formulation with $\vec\omega=(0.4,0.8,1.2)$ and $\vec l = (1,64,128,256)$.

\subsection{Analysis Details}

\paragraph{Details on Tokenization Attention Visualization.} To produce the attention maps shown in Figure~\ref{fig:attention_map}, we visualize the cross-attention weights between \DAT{} tokenized slot tokens and image patches within the dynamic decoder $\mathcal{D}_{\text{dyn}}$. Concretely, we provide the decoder with a clean video sequence and set the denoising timestep to zero, and visualize the first cross-attention layer's attention weights. The query $\mathbf{Q}$ corresponds to the patchified video latent tokens inside the decoder, and the key-value pair $(\mathbf{K}, \mathbf{V})$ corresponds to the slot tokens $\bz_{1:L} = \mathcal{E}(\bx_{t:t+\tau})$ produced by the ViViT encoder followed by the encoder-to-slot projection. For a given slot index $i$, we extract the attention weights from all decoder patches to slot $\bz_i$, reshape them back to the spatial grid, and bilinearly upsample to the original frame resolution. The resulting heatmaps are overlaid on the raw video frames. As shown in Figure~\ref{fig:attention_map}, earlier slot indices (e.g., slot 16) focus attention on regions of active motion, while later indices (e.g., slot 254) shift attention toward static background elements, confirming the dynamic-to-static ordering induced by \DAT{}. 

\paragraph{Details on Prediction Error Spatial Distribution.} To produce the error maps shown in Figure~\ref{fig:error_analysis}, we compute per-pixel absolute error between predicted and GT frames across all compared methods in the RT-1 domain. For each method, we compute the mean absolute error across color channels at every pixel, yielding a single-channel error map per frame. This pixel-wise error is mapped to an RGB heatmap and blended onto the \emph{predicted} frame, so that the spatial distribution of errors can be interpreted in the context of the generated scene content.
\clearpage

\section{Extended Experiments and Visualizations}
\label{app:ext_exp}
\subsection{Real2Sim Policy Evaluation}
\label{sec:policy-eval-details}

\paragraph{Background.}
A world model that faithfully captures environment dynamics should not only produce visually plausible rollouts but also aid downstream decision-making.  Real2Sim policy evaluation~\cite{li2025evaluating} provides a principled protocol for this purpose: by using simulation or world models as a surrogate simulator, one can assess whether it correctly distinguishes successful from unsuccessful manipulation trajectories and, crucially, whether it preserves the absolute success rate and relative performance ranking among policies observed in the real world. This interactive evaluation is particularly demanding because prediction errors compound over long horizons; a reliable world model must maintain both visual coherence and physical plausibility throughout the entire rollout.

\paragraph{Task Setup.}
We follow the evaluation protocol of~\cite{li2025evaluating} as adopted in~\cite{huang2025vid2world0}. We select the \emph{open drawer} task from the RT-1 dataset and evaluate four policies at different training stages: RT-1~(Begin), RT-1~(15\%), RT-1~(Converged)~\cite{rt_1}, and RT-1-X. For each policy, we sample $N{=}50$ trajectories and extract their initial frames. These frames are provided to the policy to produce actions, which are then fed into the world model to predict the next observation. The policy subsequently acts on the predicted frames in a closed-loop, interactive manner. We autoregressively generate each new frame conditioned on the last $L=8$ frames. This process continues until a rollout of length $H{=}40$ is reached. We then apply a trajectory-level success verifier $\psi$ to each completed rollout and compute the overall success rate as $\frac{1}{N}\sum_{n=1}^{N}\psi(\mathbf{o}_{0:H}^{(n)})$. The full procedure is described in Algorithm~\ref{alg:policy-eval}.

\paragraph{Baselines and Policy Models.}
\begin{wraptable}{r}{0.48\textwidth}
  \centering
  \vspace{-8pt}
  \caption{\textbf{Policy evaluation success rates (\%).} Each policy is evaluated in the real environment (Ground Truth) and in three world models.}
  \label{tab:policy-eval}
  \renewcommand{\arraystretch}{1.25}
  \setlength{\tabcolsep}{5pt}
  \resizebox{\linewidth}{!}{%
  \begin{tabular}{l cccc}
  \toprule
  & \textbf{RT-1}  & \textbf{RT-1}  & \textbf{RT-1}  & \textbf{RT-1-X} \\
  & \textbf{(Begin)} & \textbf{(15\%)} & \textbf{(Conv.)} & \\
  \midrule
  Ground Truth          & 0   & 70.4 & 81.5 & 51.9 \\
  \midrule
  CF (Ours)             & 2   & 54   & 64   & 40   \\
  ElasticTok            & 6   & 40   & 48   & 36   \\
  SD-VAE                & 2   & 36   & 32   & 30   \\
  \bottomrule
  \end{tabular}%
  \vspace{-12pt}
  }
\end{wraptable}
We compare CF against the same two tokenization baselines used in our main experiments: SD-VAE and ElasticTok, both equipped with the same diffusion backbone and rollout protocol. For each world model, we run the full interactive evaluation loop described above, producing 50 rollouts per policy. We evaluate four RT-1 policies that span a broad range of training progress: RT-1~(Begin), an early-stage checkpoint with near-zero real-world success; RT-1~(15\%), trained on 15\% of the data; RT-1~(Converged), the fully trained model; and RT-1-X, the cross-embodiment variant. Their real-world success rates (Table~\ref{tab:policy-eval}) increase monotonically from Begin to Converged, with RT-1-X falling between Begin and 15\%. This monotonic spread provides a clear ground-truth ranking: a faithful world model should recover this ordering in its predicted success rates.
\clearpage
\begin{algorithm}[t]
\caption{Real2Sim Policy Evaluation}\label{alg:policy-eval}
\begin{algorithmic}[1]
\REQUIRE World model $P(\mathbf{o}_{t+1}\mid\mathbf{o}_{\leq t},\,\mathbf{a}_{\leq t})$,\;
         policy $\pi(\mathbf{a}_t\mid\mathbf{o}_t)$,\;
         task $\kappa$,\;
         initial-frame set $\mathcal{D}_\kappa$,\;
         verifier $\psi(\mathbf{o}_{0:H})\!\to\!\{0,1\}$.
\STATE $\mathrm{success\_count} \leftarrow 0$
\FOR{$n = 1,\dots,N$}
    \STATE Sample initial frame $\mathbf{o}_0$ from $\mathcal{D}_\kappa$
    \FOR{$t = 0,\dots,H{-}1$}
        \STATE Sample action $\mathbf{a}_t \sim \pi(\cdot \mid \mathbf{o}_t)$
        \IF{$t < L$}
            \STATE $\mathbf{o}_{t+1} \sim P(\mathbf{o}_{t+1}\mid \mathbf{o}_{\leq t},\,\mathbf{a}_{\leq t})$
        \ELSE
            \STATE $\mathbf{o}_{t+1} \sim P(\mathbf{o}_{t+1}\mid \mathbf{o}_{t-L:t},\,\mathbf{a}_{t-L:t})$
        \ENDIF
    \ENDFOR
    \STATE $\mathrm{success\_count} \leftarrow \mathrm{success\_count} + \psi(\mathbf{o}_{0:H})$
\ENDFOR
\RETURN $\mathrm{success\_rate} = \tfrac{1}{N}\cdot\mathrm{success\_count}$
\end{algorithmic}
\end{algorithm}
\paragraph{Experimental Results.}
\begin{wrapfigure}{r}{0.48\textwidth}
  \vspace{-8pt}
  \centering
  \includegraphics[width=\linewidth]{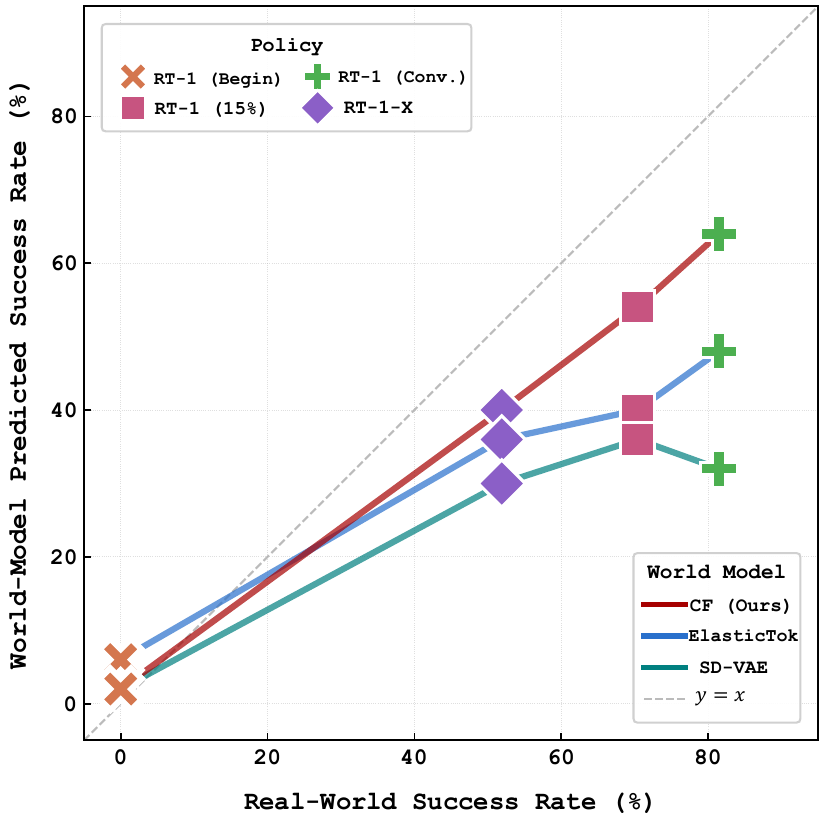}
  \caption{\textbf{Policy evaluation correlation.} Each line represents a world model; each marker denotes a policy. Points closer to the $y{=}x$ diagonal indicate more faithful policy ranking.}
  \label{fig:policy_eval}
\end{wrapfigure}
\vspace{-5pt}
We assess each world model by comparing its predicted policy success rates against the real-world ground-truth rates (Table~\ref{tab:policy-eval}). As shown in Figure~\ref{fig:policy_eval}, CF~(Ours) tracks real-world performance most accurately across all four policies, preserving both the correct ranking and approximate magnitude of success rates. In particular, SD-VAE inverts the ranking between RT-1~(15\%) and RT-1~(Converged), predicting a higher success rate for the weaker policy, whereas both ElasticTok and CF preserve the correct ordering. CF further stands out by maintaining the tightest alignment with the $y{=}x$ diagonal, indicating that its predicted success rates most closely approximate absolute real-world performance rather than merely recovering the relative ranking.

\paragraph{Human Verification.} In practice, the trajectory success verifier $\psi$ is realized through human evaluation. Annotators watch each generated rollout depicting the robot attempting to open a drawer and label the outcome according to the following criteria:
\begin{itemize}
  \item \textbf{Success}: By the final frame, the drawer is visibly pulled out from the cabinet, with a clear gap between the drawer face and the cabinet frame, and the drawer remains stationary in the open position.
  \item \textbf{Failure}: The drawer remains closed or barely displaced, the robot fails to grasp the handle, the drawer slides back after being pulled, the robot jams or stops prematurely, or the viewpoint prevents the annotator from confirming that the drawer has been opened.
\end{itemize}

\subsection{Object Tracking}
Beyond perceptual metrics, we evaluate whether generated videos preserve the physical dynamics of manipulated objects by measuring object tracking fidelity in RT-1 domain. While metrics such as FVD and LPIPS capture global perceptual similarity, they do not directly assess whether the model faithfully reproduces object motion trajectories, a property critical for downstream decision-making in robotic manipulation.
 
\paragraph{Setup.} We evaluate on 64 episodes from the RT-1 dataset. For each episode, we first obtain class-agnostic object masks on the ground-truth frame at generation onset using SAM (ViT-Base)~\cite{kirillov2023segany}, filtering by area (0.5\%--50\% of image) and deduplicating by IoU $> 0.5$. We then sample 10 query points uniformly within each detected mask and track them using CoTracker3 \citep{karaev24cotracker3} (offline mode) across 16 generated frames. The trackers starts at the first conditioned frames, and tracks object positions throughout generation. We compare our model with baselines in the main paper, with the same backbone and different tokenizers: SD-VAE and ElasticTok.
 
\paragraph{Metric.} We propose a two-component tracking error that decomposes into \emph{base} (accuracy of survived points) and \emph{penalty} (cost for lost points):
\begin{equation}
  \text{Total Error} = \underbrace{\frac{\sum_{t,i} \mathbf{1}[\text{tracked}] \cdot \|\mathbf{g}_t^i - \mathbf{p}_t^i\|}{\sum_{t,i} \mathbf{1}[v_t^i = 1]}}_{\text{Base}}
  + \underbrace{\frac{\sum_{t,i} \mathbf{1}[\text{lost}] \cdot c_t^i}{\sum_{t,i} \mathbf{1}[v_t^i = 1]}}_{\text{Penalty}},
\end{equation}
where $\mathbf{g}_t^i$ and $\mathbf{p}_t^i$ are the ground-truth and predicted positions of point $i$ at frame $t$, $v_t^i$ denotes ground-truth visibility, and $c_t^i$ is a cumulative penalty that accumulates the ground-truth frame-to-frame displacement for each frame the point remains lost:
\begin{equation}
  c_t^i = \begin{cases}
    0 & \text{if tracked at } t, \\
    c_{t-1}^i + \|\mathbf{g}_t^i - \mathbf{g}_{t-1}^i\| & \text{if lost at } t.
  \end{cases}
\end{equation}
This design ensures that the penalty is proportional to the motion importance of lost points: losing a static background point incurs negligible cost, while losing a point undergoing significant motion incurs a large, accumulating penalty.

All metrics are computed per episode and then averaged equally across episodes, including the penalty per lost point (set to zero for episodes with no lost points).
 
\begin{figure}[t]
  \centering
  \includegraphics[width=\linewidth]{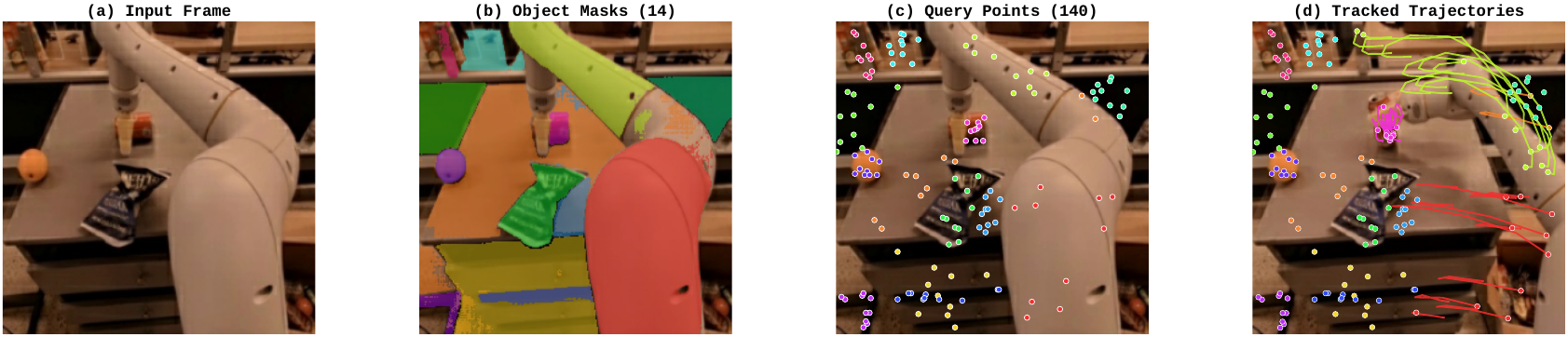}
  \caption{\textbf{Object Tracking Evaluation Overview.} 
  Given the ground-truth frame at generation onset, SAM extracts class-agnostic object masks from which query points are uniformly sampled. 
  CoTracker3 then tracks these points across both ground-truth and predicted videos.}\label{fig:tracking_overview}
\end{figure}
\begin{table}[t]
  \centering
  \caption{\textbf{Object Tracking Evaluation on RT-1.} We measure tracking fidelity across 64 episodes using CoTracker3 with SAM-initialized query points.}\label{tab:tracking}
  \setlength{\tabcolsep}{5pt}
  \begin{tabular}{lccccc}
    \toprule
    Method & Base$\downarrow$ & Penalty$\downarrow$ & \textbf{Total}$\downarrow$ & Tracked\% & Win Rate \\
    \midrule
    ElasticTok & \textbf{0.871} & 3.854 & 4.725 & 81.3\% & 6.2\% \\
    SD-VAE     & 1.473 & 2.624 & 4.097 & \textbf{91.3\%} & 45.3\% \\
    \textbf{\DAT\ (Ours)} & 1.462 & \textbf{2.451} & \textbf{3.914} & 90.8\% & \textbf{48.4\%} \\
    \bottomrule
  \end{tabular}
\end{table}
\begin{figure}[H]
  \centering
  \includegraphics[width=\linewidth]{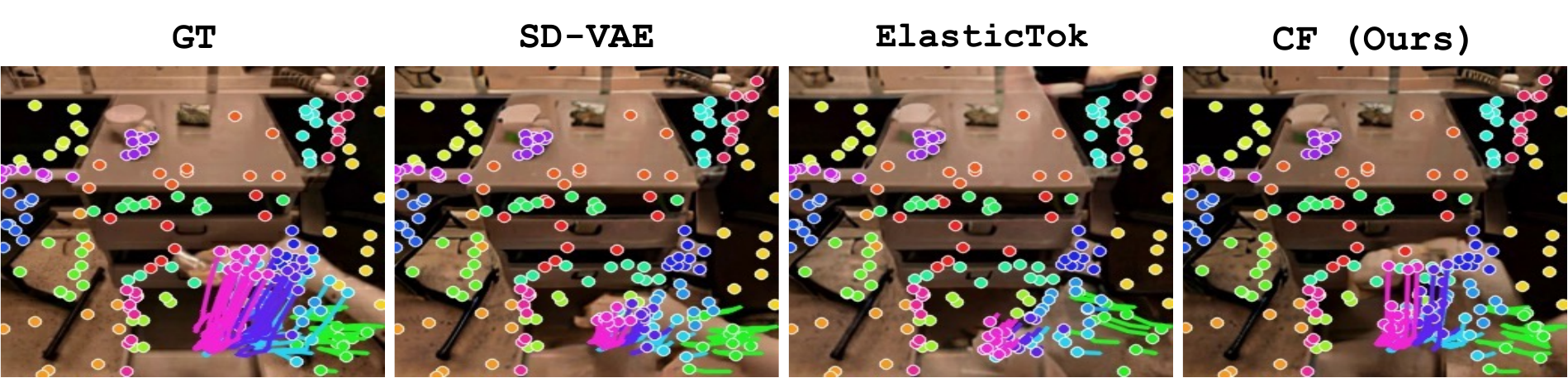}
  \caption{\textbf{Object Tracking Trajectories on RT-1.} CoTracker3 trajectories overlaid on ground-truth and predicted frames.}\label{fig:tracking_vis}
\end{figure}

\paragraph{Results.} As shown in Table~\ref{tab:tracking}, Compression Forcing achieves the the highest per-episode win rate (48.4\%). While ElasticTok achieves the lowest base error (0.871\,px), it suffers from the highest point loss rate (18.7\%), indicating poor structural consistency in generated videos. SD-VAE maintains good point survival (91.3\%) but incurs higher penalty per lost point (21.1\,px), suggesting that its lost points tend to lie in high-motion regions. In contrast, Compression Forcing balances tracking accuracy with point survival: when points are lost, they tend to be in lower-motion regions (19.7\,px per lost point), indicating that our model better preserves object dynamics precisely where it matters most, at contact points and regions of active manipulation. This finding is consistent with the per-pixel error analysis in Figure~\ref{fig:error_analysis}, where Compression Forcing's error accumulation is less pronounced at dynamic contact regions compared with baselines.

\clearpage
\subsection{Generation Results on Kinetics}
We present extended qualitative results on the Kinetics-600 video prediction domain. All rollouts are conditioned on 16 context frames and autoregressively generate 8 predicted frames across diverse motion-rich natural videos.
\begin{figure}[H]
    \centering
    \includegraphics[width=\linewidth]{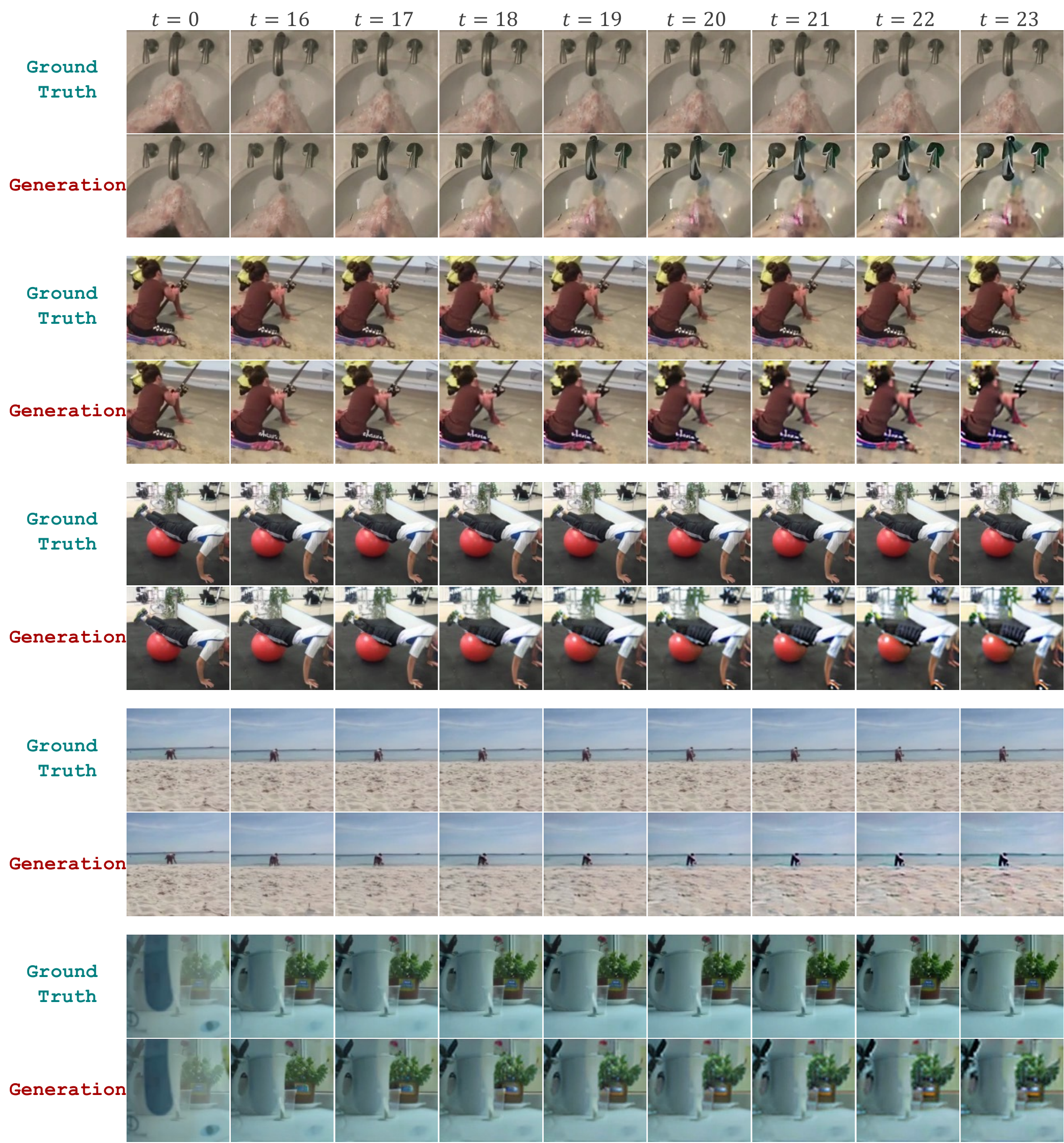}
    \caption{\textbf{Extended generation results on Kinetics-600.} Comparison between ground truth and generated videos across diverse natural motion sequences. Conditioned on 16 frames as context, autoregressive generation.}
    \label{fig:full_kinetics_vis}
\end{figure}
As shown in Figure \ref{fig:full_kinetics_vis}, Compression Forcing produces rollouts that preserve both the structure of the scene and the continuity of motion, maintaining coherent body poses and object interactions throughout the prediction horizon.

\clearpage
\subsection{Generation Results on CS:GO}

We present extended qualitative results of Compression Forcing on the CS:GO environment. All rollouts are conditioned on 4 context frames and autoregressively generate 8 predicted frames.

\begin{figure}[h]
    \centering
    \includegraphics[width=\linewidth]{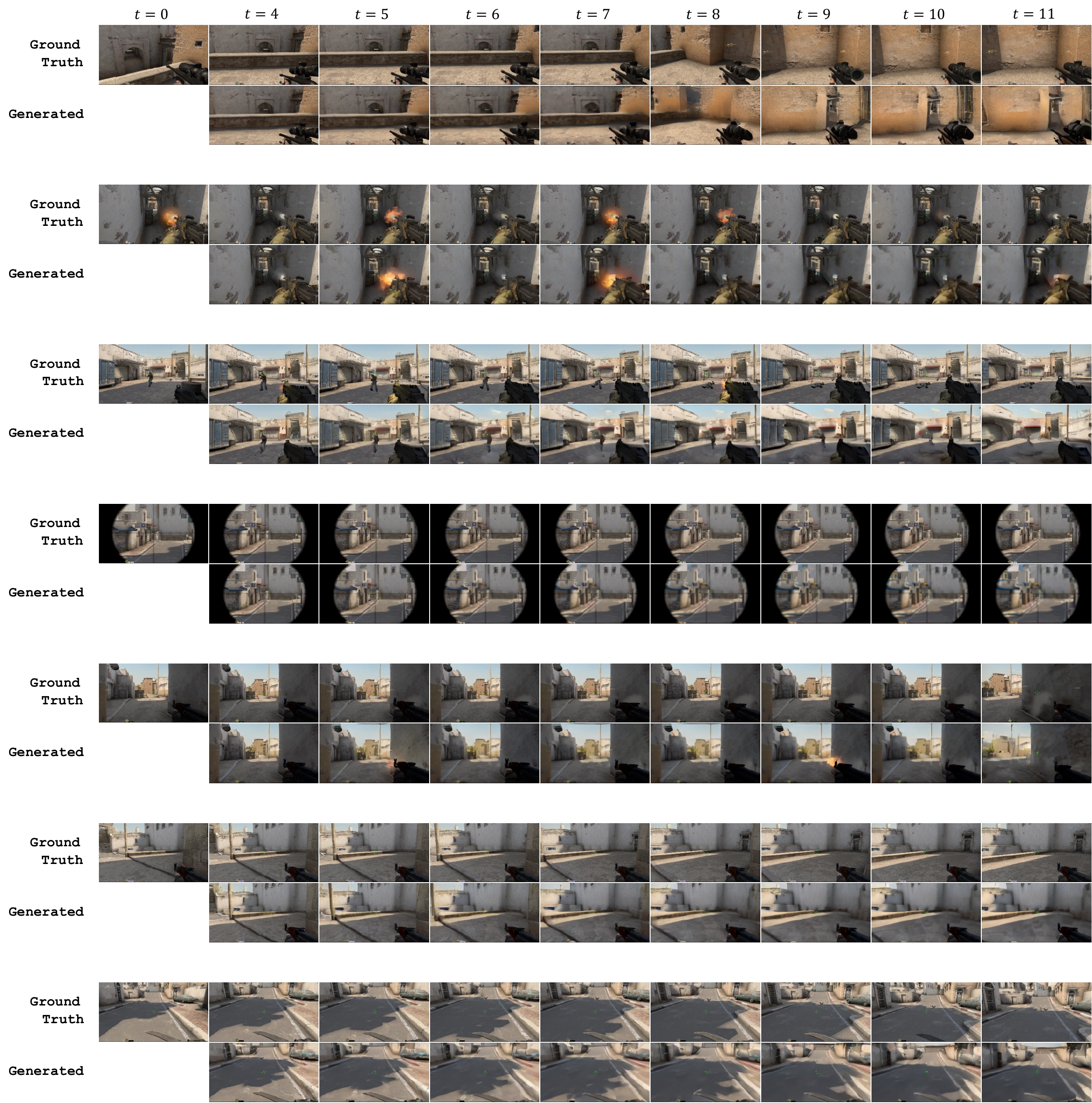}
    \caption{\textbf{Extended generation results on CS:GO.} Comparison between ground truth and generated videos across diverse gameplay scenarios. The first four frames are provided as context; subsequent frames are autoregressively generated.}
    \label{fig:full_csgo_vis}
\end{figure}

\paragraph{General Generation Quality.} Figure \ref{fig:full_csgo_vis} showcases diverse gameplay scenarios including firefights, aiming through lenses, and open-area traversal. Across all examples, Compression Forcing produces temporally coherent rollouts that preserve scene geometry and maintain consistent weapon positioning. Even under viewpoint changes, the model faithfully renders environmental structures such as walls, crates, and doorways without structural drift.

\begin{figure}[t]
    \centering
    \includegraphics[width=\linewidth]{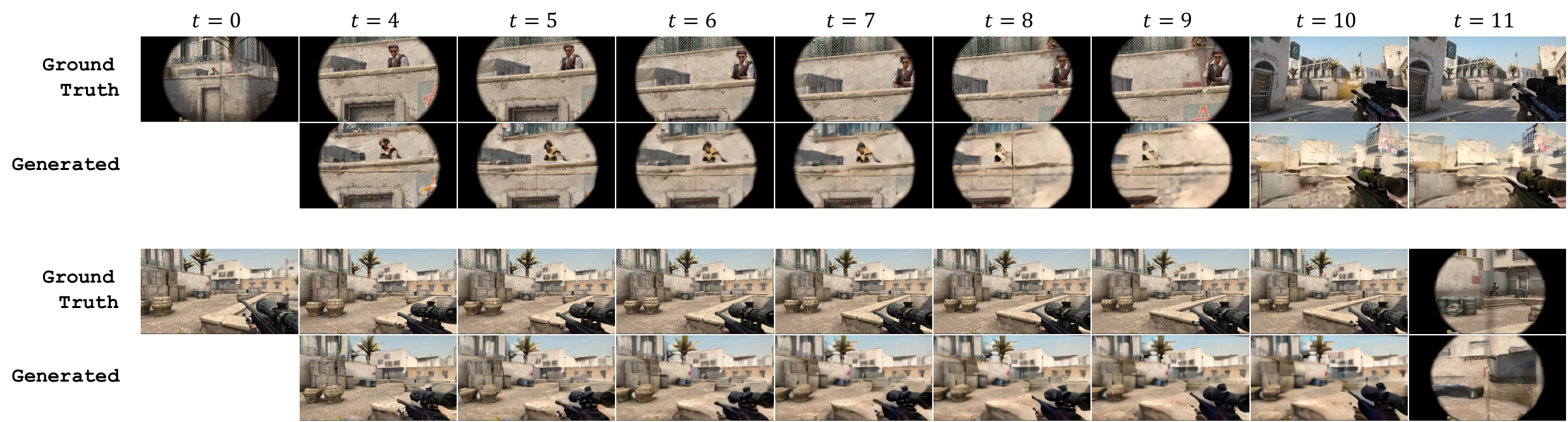}
    \caption{\textbf{Action following: scope transitions in CS:GO.} Top: the player exits the scope view; the model correctly captures the transition from zoomed-in to standard field of view. Bottom: the player enters scope mode; the generated frames accurately reflect the visual change. Zoom in for details.}
    \label{fig:csgo_action_follow}
\end{figure}

\paragraph{Action Following.} Figure \ref{fig:csgo_action_follow} illustrates the model's fidelity to action-conditioned transitions. In the first example, the player exits the scope view, and the generated frames correctly capture the transition from the zoomed-in scope overlay to the standard field of view. In the second example, the player enters scope mode, and Compression Forcing accurately renders the narrowing of the visual field and the appearance of the scope reticle. Both examples confirm that the model has learned to associate discrete action inputs with their corresponding visual effects.

\begin{figure}[t]
    \centering
    \includegraphics[width=\linewidth]{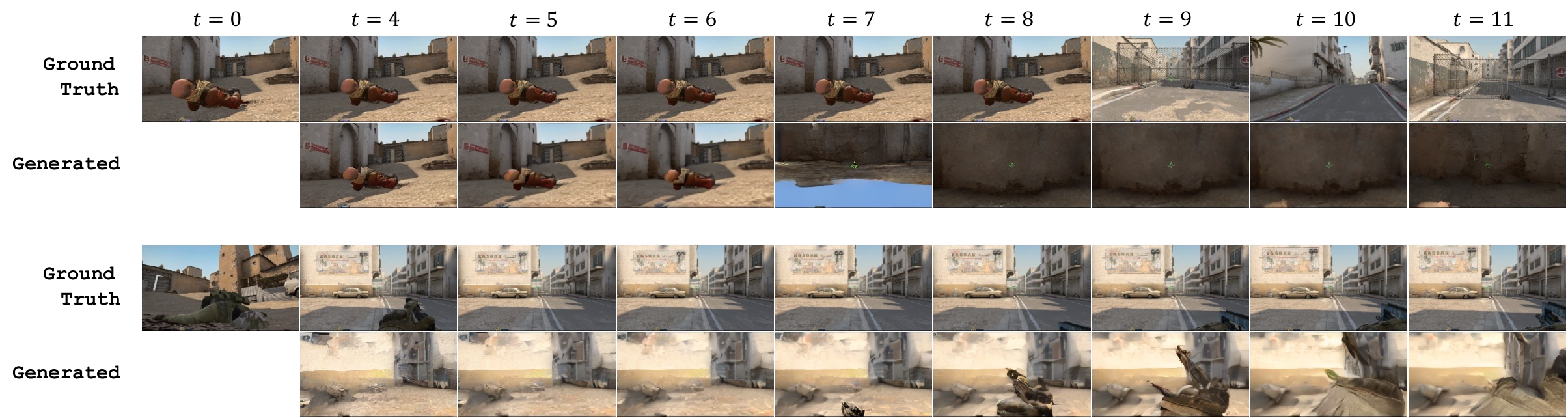}
    \caption{\textbf{Environment stochasticity: player respawn in CS:GO.} Upon death, CS:GO teleports the agent to a random spawn location. In both examples, the model generates a plausible but incorrect post-respawn scene, reflecting the inherent stochasticity of the underlying environment. Zoom in for details}
    \label{fig:csgo_sde}
\end{figure}

\paragraph{Environment Stochasticity.} Figure \ref{fig:csgo_sde} illustrates Compression Forcing's capability of capturing environment stochasticity, generating feasible yet different plausible respawn positions.

\clearpage
\subsection{Generation Results on RT-1}
\label{sec:full_rt1_vis}
We present extended qualitative results on the RT-1 robot manipulation domain. All rollouts are conditioned on 8 context frames and autoregressively generate 16 predicted frames across diverse manipulation tasks including picking, placing, and drawer interactions.

\begin{figure}[h]
    \centering
    \includegraphics[width=\linewidth]{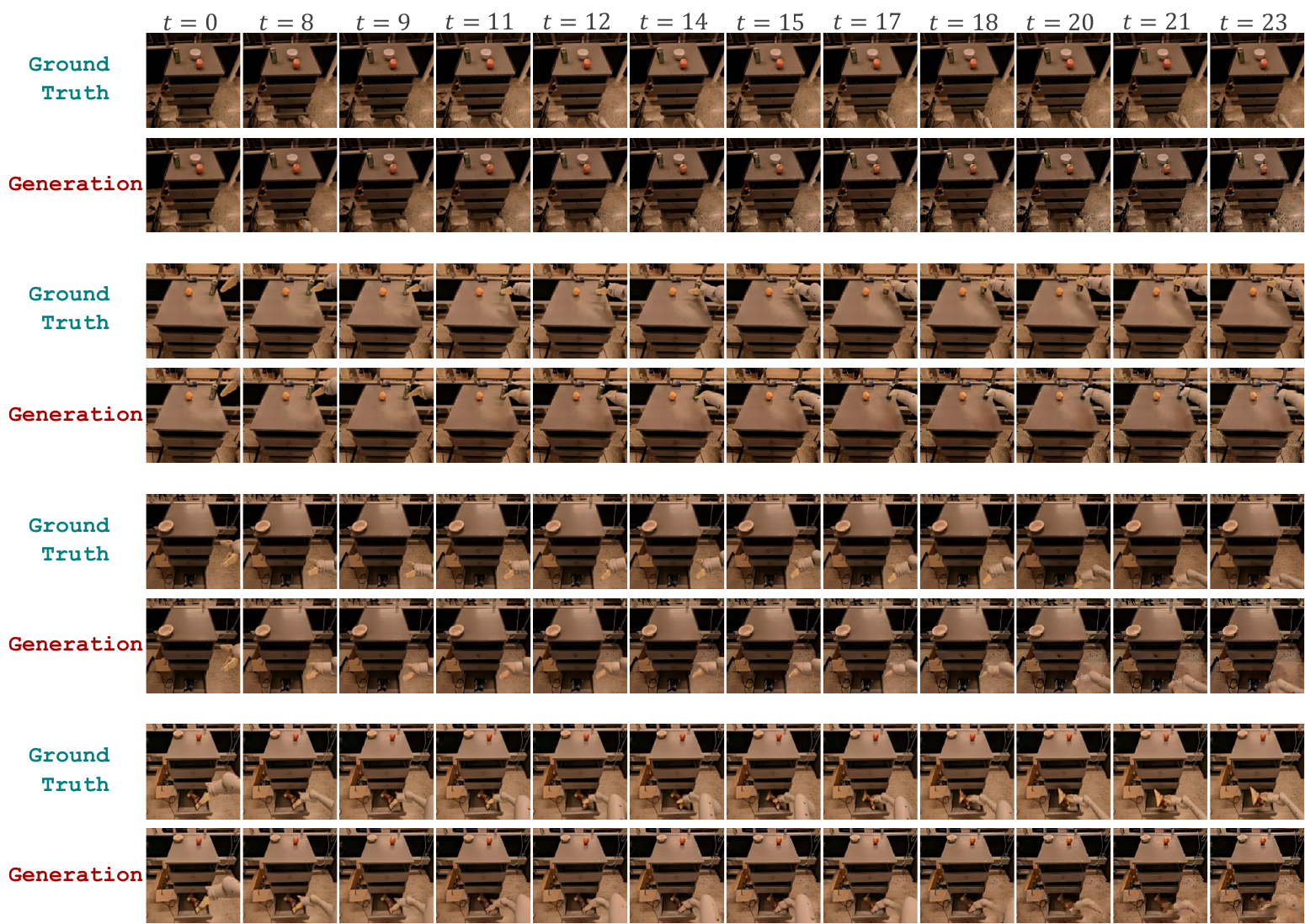}
    \caption{\textbf{Extended generation results on RT-1.} Comparison between ground truth and generated videos across diverse manipulation tasks. The first eight frames are provided as context; subsequent frames are autoregressively generated. Zoom in for details.}
    \label{fig:full_rt1_vis}
\end{figure}

Figure \ref{fig:full_rt1_vis} presents extended qualitative results on the RT-1 domain. Compression Forcing produces rollouts that accurately track the robot arm's trajectory and preserve fine-grained contact interactions with the manipulated objects throughout the prediction horizon.

\subsection{Extended Visualizations on Analysis Experiments}
\label{sec:full_analysis}
\begin{figure}[H]
    \centering
    \includegraphics[width=\linewidth]{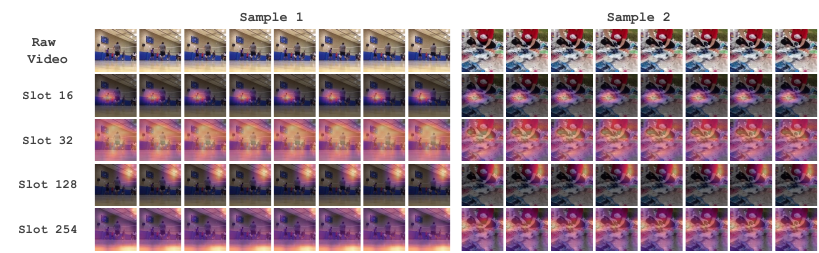}
    \caption{\textbf{Extended Tokenization Attention Maps.} Cross-attention maps between \DAT{} slot tokens and image patches in the dynamic decoder across additional samples. Consistent with Figure \ref{fig:attention_map}, earlier slots attend to regions of active motion, while later slots progressively shift toward static background elements, confirming the generality of \DAT's dynamics-to-static ordering.}
    \label{fig:full_attention_map}
\end{figure}

\begin{figure}[H]
    \centering
    \includegraphics[width=\linewidth]{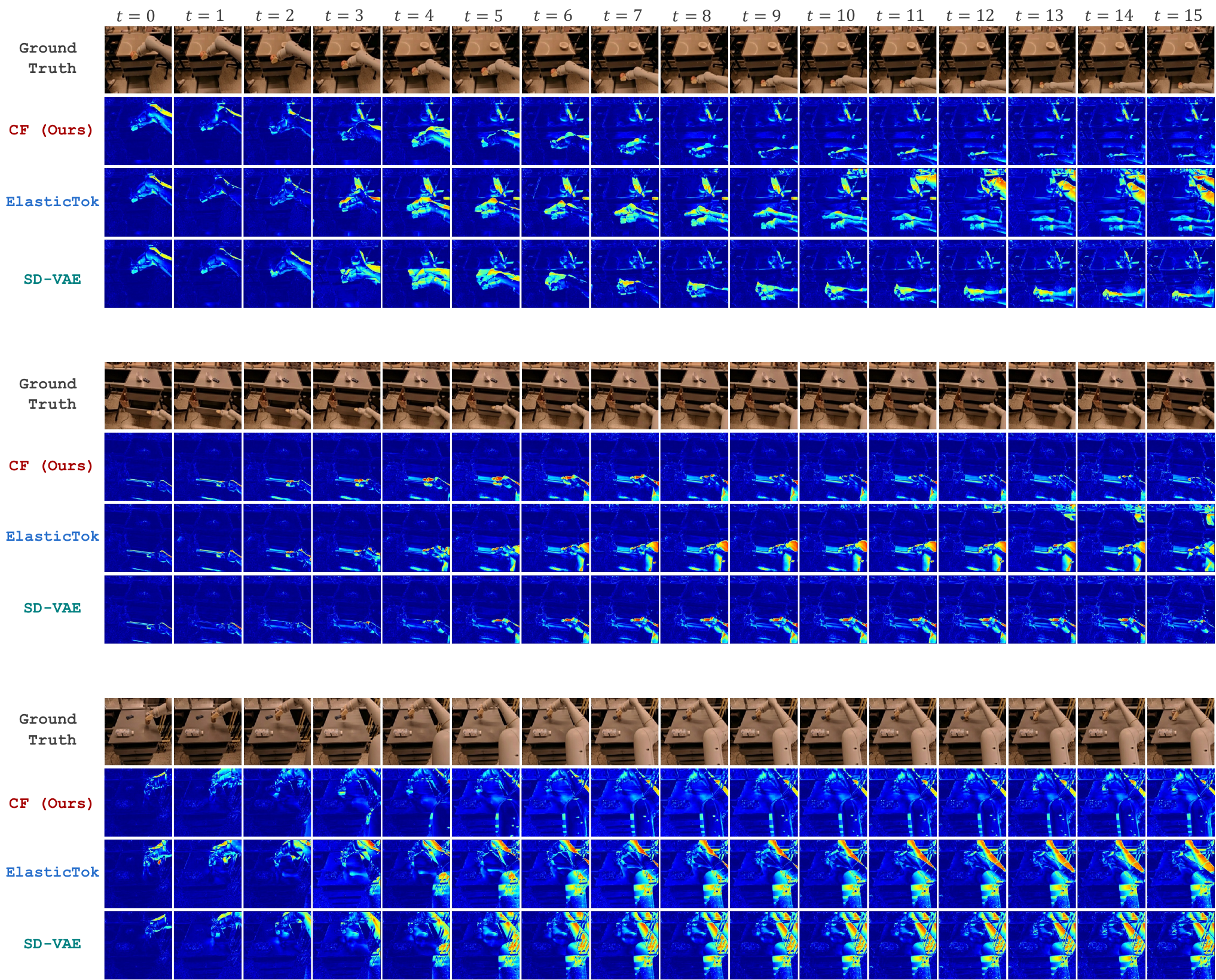}
    \caption{\textbf{Extended Visualizations of Pixel-wise Prediction Error}. Compression Forcing induces less error compared to both baselines (ElasticTok and SD-VAE), specially around regions of dynamic contact.}
    \label{fig:full_error_map}
\end{figure}

We provide extended visualizations for the two main analysis experiments discussed in Section \ref{subsec:analysis}. Figure \ref{fig:full_attention_map} presents additional attention map samples from \DAT{} across different video sequences, and Figure \ref{fig:full_error_map} extends the per-pixel prediction error comparison to additional RT-1 episodes.

\end{document}